%% file: main.tex
\documentclass[10pt,twocolumn,letterpaper]{article}

\usepackage{cvpr}              

\input{preamble}

\definecolor{cvprblue}{rgb}{0.21,0.49,0.74}
\usepackage[pagebackref,breaklinks,colorlinks,citecolor=cvprblue]{hyperref}
\usepackage{times}
\usepackage{epsfig}
\usepackage{graphicx}
\usepackage{amsmath}
\usepackage{amssymb}
\usepackage{pifont}
\usepackage{booktabs}
\usepackage{arydshln}
\usepackage{color}
\usepackage{colortbl}
\usepackage{subcaption}
\usepackage{multirow}
\usepackage{float}
\usepackage{rotfloat}
\usepackage{capt-of}
\usepackage{longtable}
\usepackage{diagbox}
\usepackage{makecell}
\usepackage{pgfplots}
\usepackage{xspace}
\usepackage{wrapfig}
\usepackage{enumitem}

\usepackage[accsupp]{axessibility}
\def\paperID{15774} 
\def\confName{CVPR}
\def\confYear{2024}

\newcommand{\cmark}{\scalebox{0.8}{\ding{52}}}
\newcommand{\xmark}{\ding{55}}  

\usepackage[capitalize]{cleveref}
\crefname{section}{Sec.}{Secs.}
\Crefname{section}{Section}{Sections}
\Crefname{table}{Table}{Tables}
\crefname{table}{Tab.}{Tabs.}
\newcommand{\ourmodel}{DetCLIPv3\xspace}
\newcommand{\ourmodelft}{DetCLIPv3+FT\xspace}
\newcommand{\ourdata}{GranuCap50M\xspace}
\newcommand{\ourdatasub}{GranuCap600K\xspace}

\title{\vspace{-1.5em}DetCLIPv3: Towards Versatile Generative Open-vocabulary Object Detection\vspace{-0.5em} }

\author{%
    Lewei Yao$^{1,2}$, Renjie Pi$^{1}$, Jianhua Han$^{2}$, Xiaodan Liang$^{3}$, \\Hang Xu$^{2}$$^\dagger$, Wei Zhang$^{2}$, Zhenguo Li$^{2}$, Dan Xu$^{1}$$^\dagger$ \\
  \small{$^1$Hong Kong University of Science and Technology, $^2$Huawei Noah's Ark Lab} \\
  \small {$^3$Shenzhen Campus of Sun Yat-Sen University }
}
\begin{document}

\twocolumn[{
\vspace{-2em}
\maketitle
\vspace{-1em}

\begin{center}
    \centering 
    \vspace{-2em}
    \includegraphics[width=1\linewidth]{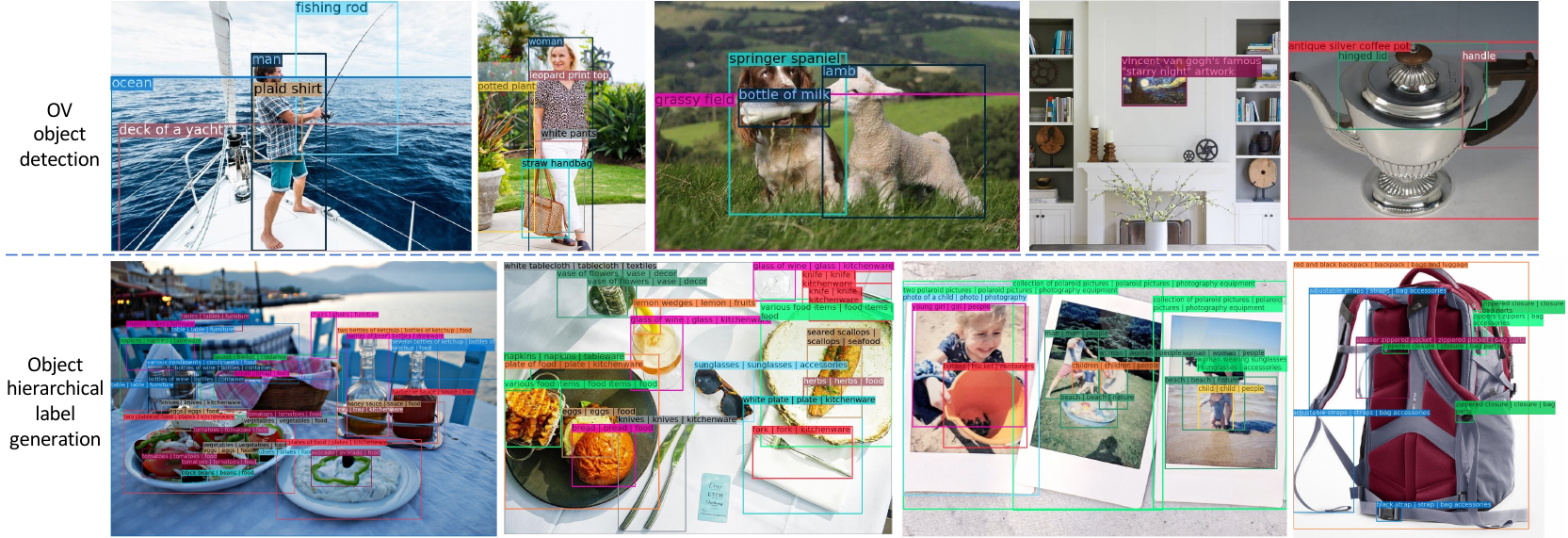}
    \vspace{-2em}
    \captionof{figure}{The versatility of \ourmodel supports both open-vocabulary object detection (OVD) and the generation of hierarchical object labels. \textbf{Top}: when provided with extracted noun phrases from image-text pair captions as input, \ourmodel can detect a broad spectrum of visual concepts. \textbf{Bottom}: In the absence of predefined categories as input, DetCLIPv3 detects potential objects and generates multi-granularity hierarchical labels for them, formatted as '\textit{phrase $|$ category $|$ parent category}'. \ourmodel offers a more comprehensive interpretation of objects, significantly expanding the application scope of OVD systems. Zoom in for the best viewing.}
    \label{fig:det_gen_results}
\end{center}
}]

\let\thefootnote\relax\footnotetext{$^\dagger$Corresponding author: xu.hang@huawei.com, danxu@cse.ust.hk }

\input{sec/abstract}    

\input{sec/intro}
\input{sec/related_work}
\input{sec/method}
\input{sec/experiments}
\input{sec/conclusion}

\input{sec/appendix}

{
    \small
    \bibliographystyle{ieeenat_fullname}
    \bibliography{main}
}


\end{document}

%% file: preamble.tex
\usepackage[dvipsnames]{xcolor}
\definecolor{lightblue}{HTML}{E8F0FE}
\definecolor{gray}{HTML}{9aa0a6}
\definecolor{lightpink}{HTML}{F48FB1}
\definecolor{lightred}{HTML}{EF9A9A}
\definecolor{lightcyan}{HTML}{80DEEA}

\definecolor{lightred}{rgb}{1.0, 0.4, 0.4}
\definecolor{textlightblue}{rgb}{0.4, 0.698, 1.0}
\definecolor{deeppurple}{rgb}{0.294, 0.0, 0.51}
\definecolor{navyblue}{rgb}{0.18, 0.45, 0.727}
\definecolor{forestgreen}{rgb}{0.133, 0.545, 0.133}

\newcommand{\red}[1]{\textcolor{red}{#1}}
\newcommand{\lightred}[1]{\textcolor{lightred}{#1}}
\newcommand{\lightblue}[1]{\textcolor{textlightblue}{#1}}
\newcommand{\deeppurple}[1]{\textcolor{deeppurple}{#1}}
\newcommand{\navyblue}[1]{\textcolor{navyblue}{#1}}
\newcommand{\forestgreen}[1]{\textcolor{forestgreen}{#1}}
\newcommand{\orange}[1]{\textcolor{orange}{#1}}
\newcommand{\gray}[1]{\textcolor{gray}{#1}}

\newcommand{\darkgreen}[1]{\textcolor{green!50!black}{#1}}

\renewcommand{\paragraph}[1]{\vspace{1.1mm}\noindent\textbf{#1}}
\newcommand{\tablestyle}[2]{\setlength{\tabcolsep}{#1}\renewcommand{\arraystretch}{#2}\centering\footnotesize}
\makeatletter
\DeclareRobustCommand\onedot{\futurelet\@let@token\@onedot}
\def\@onedot{\ifx\@let@token.\else.\null\fi\xspace}
\def\eg{\textit{e.g}\onedot}
\def\Eg{\textit{E.g}\onedot}
\def\ie{\textit{i.e}\onedot}

\makeatother


%% file: sec/abstract.tex
\begin{abstract}
\vspace{-1em}
Existing open-vocabulary object detectors typically require a predefined set of categories from users, significantly confining their application scenarios. 
In this paper, we introduce \ourmodel, a high-performing detector that excels not only at both open-vocabulary object detection, but also generating hierarchical labels for detected objects. \ourmodel is characterized by three core designs: 1. Versatile model architecture: we derive a robust open-set detection framework which is further empowered with generation ability via the integration of a caption head. 2. High information density data: we develop an auto-annotation pipeline leveraging visual large language model to refine captions for large-scale image-text pairs, providing rich, multi-granular object labels to enhance the training.  3. Efficient training strategy: we employ a pre-training stage with low-resolution inputs that enables the object captioner to efficiently learn a broad spectrum of visual concepts from extensive image-text paired data. This is followed by  a fine-tuning stage that leverages a small number of high-resolution samples to further enhance detection performance. With these effective designs, \ourmodel demonstrates superior open-vocabulary detection performance, \eg, our Swin-T backbone model achieves a notable 47.0 zero-shot fixed AP on the LVIS minival benchmark, outperforming GLIPv2, GroundingDINO, and DetCLIPv2 by 18.0/19.6/6.6 AP, respectively. \ourmodel also achieves a state-of-the-art 19.7 AP in dense captioning task on VG dataset, showcasing its strong generative capability.

\end{abstract}

%% file: sec/intro.tex
\vspace{-1.5em} 
\section{Introduction}
\label{sec:intro}

Recent progress in open-vocabulary object detection (OVD) has achieved the ability to identify and localize a diverse range of objects~\citep{gao2021towards,glip, fontanel2022detecting,inkawhich2022self, detclip, detclipv2, xdetr, zhang2023simple, grounddino}. However, these models are limited by their reliance on a predefined object category list during inference, which hinders their usage in practical scenarios.

\begin{figure}
    \centering
    \includegraphics[width=0.47\textwidth]{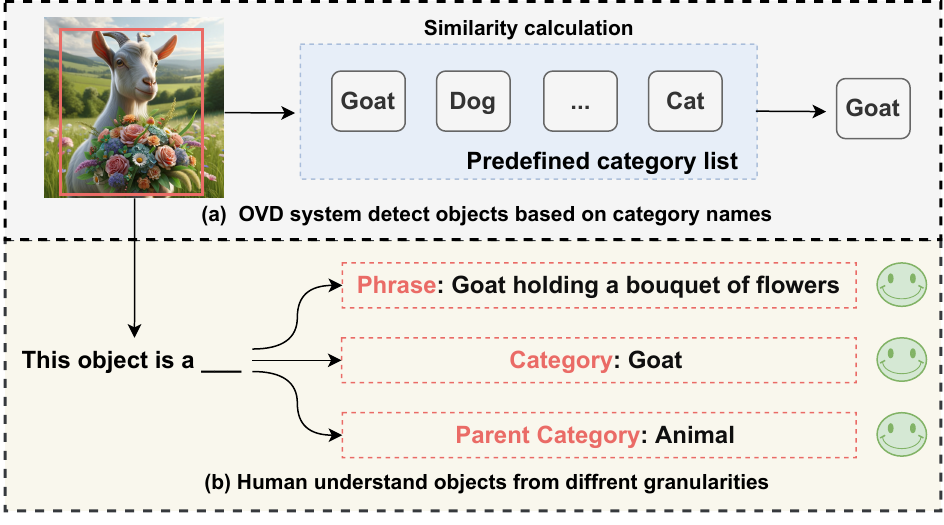}
    \vspace{-1em}
    \caption{(a): Existing open-vocabulary object detectors recognize objects based on category names; (b): Humans interpret visual concepts from multiple hierarchies and granularities.}
    \vspace{-1.5em}
    \label{fig:intro}
\end{figure}

In contrast to current open-vocabulary object detection (OVD) methods that recognizes objects solely based on category names, human cognition demonstrates much more versatility. As illustrated in Figure~\ref{fig:intro}, humans are able to understand objects from different granularities, in a hierarchical manner. This multi-level recognition ability showcases the rich visual understanding that humans possess, which is yet to be achieved in contemporary OVD systems.

To address the above limitations, we introduce \ourmodel, a novel object detector that enhances the scope of open-vocabulary object detection. \ourmodel is able to not only recognize objects based on provided category names but also generate hierarchical labels for each detected object. This feature offers two advantages: 1) owing to the superior generative ability, the detector is applicable even in the absence of the appropriate input object category; 2) the model is able to provide a comprehensive and hierarchical description about the objects, rather than simply recognizing them based on given categories. Specifically, \ourmodel is characterized by three core designs:

\textbf{Versatile model architecture:} \ourmodel is grounded on a robust open-vocabulary (OV) detector, which is further empowered with a object captioner to provide generative capabilities. Specifically, the object captioner leverages the foreground proposals provided by the OV detector and is trained to generate hierarchical labels for each detected object through a language modeling training objective. This design allows for not only accurate localization, but also detailed descriptions of the visual concepts, and thereby offering a richer interpretation of the visual contents.

\textbf{High information density data:} The development of strong generative ability necessitates abundant training data enriched with detailed object-level descriptions. The scarcity of such comprehensive datasets (\eg, Visual Genome~\cite{vg}) presents a substantial obstacle in training effective object captioners. On the other hand, while large-scale image-text pair data are plentiful, they lack fine-grained annotations for each object. To benefit from such data, we design an auto-labeling pipeline leveraging state-of-the-art vision large language models~\cite{llava,instructblip}, which is able to provide refined image captions containing rich hierarchical object labels. With this pipeline, we derive a large-scale dataset (termed as \textbf{GranuCap50M}) for bolstering \ourmodel’s abilities in both detection and generation.

\textbf{Efficient multi-stage training:} The prohibitive training costs associated with high-resolution inputs for object detectors present a significant barrier to learning from extensive image-text pairs. To address the issue, we propose an efficient multi-stage alignment training strategy. This method initially harnesses knowledge from large-scale, low-resolution image-text datasets, followed by fine-tuning on high-quality, fine-grained, high-resolution data. The approach ensures comprehensive visual concept learning while maintaining manageable training demands.

With the effective designs, \ourmodel achieves outstanding detection and object-level generation capabilities, \eg, with a Swin-T backbone, it achieves a remarkable 47.0 zero-shot \textit{fixed} AP~\cite{fixap} on the LVIS minival benchmark, significantly outperforming predecessors like GLIPv2~\cite{glipv2}, DetCLIPv2~\cite{detclipv2}, and GroundingDINO~\cite{grounddino}. Besides, it achieves 18.4 mAP on dense captioning task, surpassing the previous SOTA method GRiT~\cite{grit} by 2.9 mAP. Extensive experiments further demonstrate the superior domain generalization and downstream transferbility of \ourmodel.

%% file: sec/related_work.tex
\vspace{-.5em}
\section{Related works}
\vspace{-.6em}
\paragraph{Open-vocabulary object detection.} Recent advancements in Open-vocabulary Object Detection (OVD) allow for identifying objects across unlimited range of categories, as seen in~\citep{zang2022open,xie2021zsd,gu2021open,regionclip,vild}. These approaches achieve OVD by incorporating pre-trained VL models like CLIP~\cite{clip} into the detector. Alternatively, expanding the detection training dataset has shown promise~\citep{mdetr, glip,glipv2,detclip,detclipv2,grounddino,lin2022learning,detic}, which combine the datasets from various task, such as classification and visual grounding.  Moreover, pseudo labeling has emerged as another effective strategy for augmenting training datasets, as demonstrated in ~\cite{regionclip, detclip, glip, scaleovdet, gao2022open, zhao2022exploiting}. However, previous OVD methods still require a predefined object categories for detection, limiting their applicability in diverse scenarios. In contrast, our \ourmodel is capable of generating rich hierarchical object labels even in the absence of category names.

\paragraph{Dense captioning.} Dense captioning aims at generating descriptions for specific image areas~\citep{densecap, li2019visualbert, CAGNet, TDC, imgG}. Recently, CapDet~\cite{capdet} and GRiT~\cite{grit} both equip the object detector with generative ability by introducing a captioner. However, they are only able to generate descriptions for limited visual concepts due to the scarcity of training data contained in \ie, Visual Genome~\cite{vg}. In contrast, we harness the rich knowledge in the large-scale image-text pairs, and enable the model to generate hierarchical label information for much broader spectrum of concepts.

\paragraph {Re-captioning for image-text paris.} Recent studies~\cite{pixart,blip,capsfusion,dalle3} highlight the issues present in current image-text pair data and have shown that recaptioned high-quality image-text pairs can significantly enhance the learning efficiency of various visual tasks, such as text-to-image generation~\cite{dalle3,pixart}, image-text retrieval~\cite{blip,blip2} and image captioning~\cite{blip,capsfusion}. We extend this idea to open-vocabulary object detection and explore how to effectively utilize the object entity information contained in image-text pairs.

%% file: sec/method.tex
\begin{figure*}
\vspace{-1.5em}
\includegraphics[width=1.0\textwidth]{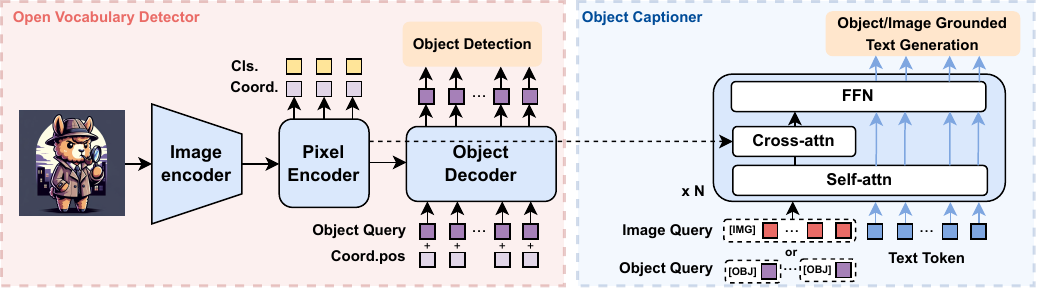} 
\vspace{-1.5em}
\caption{ The illustration for \ourmodel framework. \lightred{Left}: the OV detector is responsible for localizing objects given category names, as well as providing object proposals for the object captioner. \lightblue{Right}: The object captioner is designed to generate hierarchical labels for detected objects and also learns to generate image-level descriptions as an aid to its training.
}\label{fig:framework}
\vspace{-1em}

\end{figure*}
\vspace{-.5em}
\section{Method}
\label{sec:method}
In this section, we introduce the core designs of \ourmodel, which includes: (1) Model Architecture (Sec.~\ref{subsec:model}) - elucidating how our model enables both open-vocabulary object detection and generation of object descriptions; (2) Auto-Annotation Data Pipeline (Sec.~\ref{subsec:dataset_construct}) - detailing our approach for curating large-scale, high-quality image-text pairs, encompassing objects information across a spectrum of granularities; and (3) Training Strategy (Sec.~\ref{subsec:train strategy}) - outlining how we effectively leverage large-scale image-text datasets to facilitate object concept generation, which in turn boosts open-vocabulary detection capabilities.

\subsection{Model Design}
Figure~\ref{fig:framework} illustrates the overall framework of \ourmodel. In essence, the model is grounded on a powerful open-vocabulary detection object detector, and is equipped with an object captioner dedicated to generating hierarchical and descriptive object concepts. The model is able to function under two modes: 1) When a pre-defined category vocabularies list is provided, \ourmodel predicts the localization of objects that are mentioned in the list; 2) in the absence of the vocabulary list,  \ourmodel is able to localize the objects and generate hierachical for each of them.

\label{subsec:model}
\paragraph{Data formulation.} 
\ourmodel's training leverages datasets from multiple sources, including detection~\cite{objects365,v3det}, grounding~\cite{mdetr}, and image-text pairs~\cite{cc3m,cc12m,yfcc100m,laion400m} with bounding-box pseudo-labels (as detailed in Sec.~\ref{subsec:dataset_construct}). Following DetCLIPv1/v2~\cite{detclip,detclipv2}, we employ a \textit{parallel formulation} to unify text inputs from various data sources into a uniform format. Specifically, each input sample is structured as a triplet, $(x, \{\textbf{b}_i\}_{i=1}^N, y_{i=1}^M )$, where $x \in \mathbb{R}^{3\times H \times W}$ is the input image, $\{\textbf{b}_i| \textbf{b}_i\in \mathbb{R}^4\}_{i=1}^N$ represents a set of bounding boxes, and $y_{i=1}^M$ denotes a set of concept texts, comprising both positive and negative concepts.

For detection data, $y_{j}$ comprises class names along with their definitions (as in ~\cite{detclip,detclipv2}), applicable in both training and testing phases. The negative concepts are sampled from categories within the dataset. For grounding and image-text pair data, the positive concepts are object descriptions, while the negatives are sampled from a large-scale noun corpus (details in Sec.~\ref{subsec:dataset_construct}). During training, to increase the number of negative concepts, we collect them across all training nodes and implement a deduplication process.

\paragraph{Open vocabulary detector.}
We present a compact yet powerful detector architecture for \ourmodel, which is depicted within the red box of Figure~\ref{fig:framework}. Specifically, it is a dual-path model comprising a visual object detector $\Phi_v$ and a text encoder $\Phi_t$. The visual object detector employs a transformer-based detection architecture~\cite{detr,deformdetr,dino}, composed of a backbone, a pixel encoder, and an object decoder. The backbone and pixel encoder are responsible for extracting visual features, conducting fine-grained feature fusion, and proposing candidate object queries for the decoder. Similar to GroundingDINO~\cite{grounddino},  we utilize text features to select the top-k pixel features based on similarity, and later using their coordinate predictions to initialize the positional part of the decoder object query. However, distinctively, we abandon the computationally intensive cross-modal fusion modules designed in~\cite{grounddino}. Following previous DETR-like detectors~\cite{detr,deformdetr,dino}, our training loss is composed of three components: $\mathcal{L}_{det}=\mathcal{L}_{align}+\mathcal{L}_{box}+\mathcal{L}_{iou}$, where $\mathcal{L}_{align}$ is a contrastive focal loss~\cite{focal_loss} between regional visual features and textual concepts, while $\mathcal{L}_{box}$ and $\mathcal{L}_{iou}$ are the L1 loss and GIOU~\cite{giou} loss, respectively. To boost the performance, auxiliary losses are employed at each layer of the decoder and at the output of the encoder.

\paragraph{Object captioner.} The object captioner empowers DetCLIPV3 to generate detailed and hierarchical label for objects. To acquire the rich knowledge contained in image-text pairs, we further incorporate image-level captioning objective during training to enhance the generation capability. As illustrated in the blue box of Figure~\ref{fig:framework}, the design of the object captioner is inspired by Qformer~\cite{blip2}. Specifically, it adopts a multi-modal Transformer-based architecture with its cross-attention layer replaced by deformable attention~\cite{deformdetr} tailored for dense prediction task. The captioner’s input comprises both visual (object or image) queries and text tokens. The visual queries interact with features from the pixel encoder via cross-attention, while the self-attention layers and FFN layers are shared across different modalities. Furthermore, a multimodal causal self-attention mask~\cite{uniLM,blip2} is adopted to control the interaction between visual queries and text tokens. The training of the captioner is guided by the conventional language modeling loss $\mathcal{L}_{lm}$, with distinct input formats for object-level and image-level generation:
\begin{itemize}
    \item \textbf{Object-Level generation}. The object query and the reference points required for the deformable cross-attention are derived from the final layer output of the object decoder. The input is structured as: $<object query,\texttt{[OBJ]},text>$, where \texttt{[OBJ]} is a special task token indicating the object generation task. During training, we compute the loss using positive queries matching the ground truth. 
    During inference, to obtain foreground proposals, we select the top-k candidate object queries based on their similarity to the most frequent 15K noun concepts from our curated noun corpus (Sec.~\ref{subsec:dataset_construct}). After generating hierarchical labels for these objects, we re-calibrate their objectness scores, using the OV detector to calculate the similarity between object queries and their generated `phrase' and `category' fields. The higher of these 2 similarities is then adopted as objectness score.
    \item \textbf{Image-Level generation}. Inspired by Qformer~\cite{blip2}, we initialize 32 learnable image queries and use a set of fixed reference points. Specifically, we sample 32 locations from equal intervals from the reference points of the pixel encoder.  Similar to object-level generation, the input is structured as  $<image query,\texttt{[IMG]},text>$, with \texttt{[IMG]} being a special task token indicating image generation. The inference process of image-level generation is consistent with the training.
\end{itemize}

\begin{figure*}
\vspace{-1em}
\includegraphics[width=1.0\textwidth]{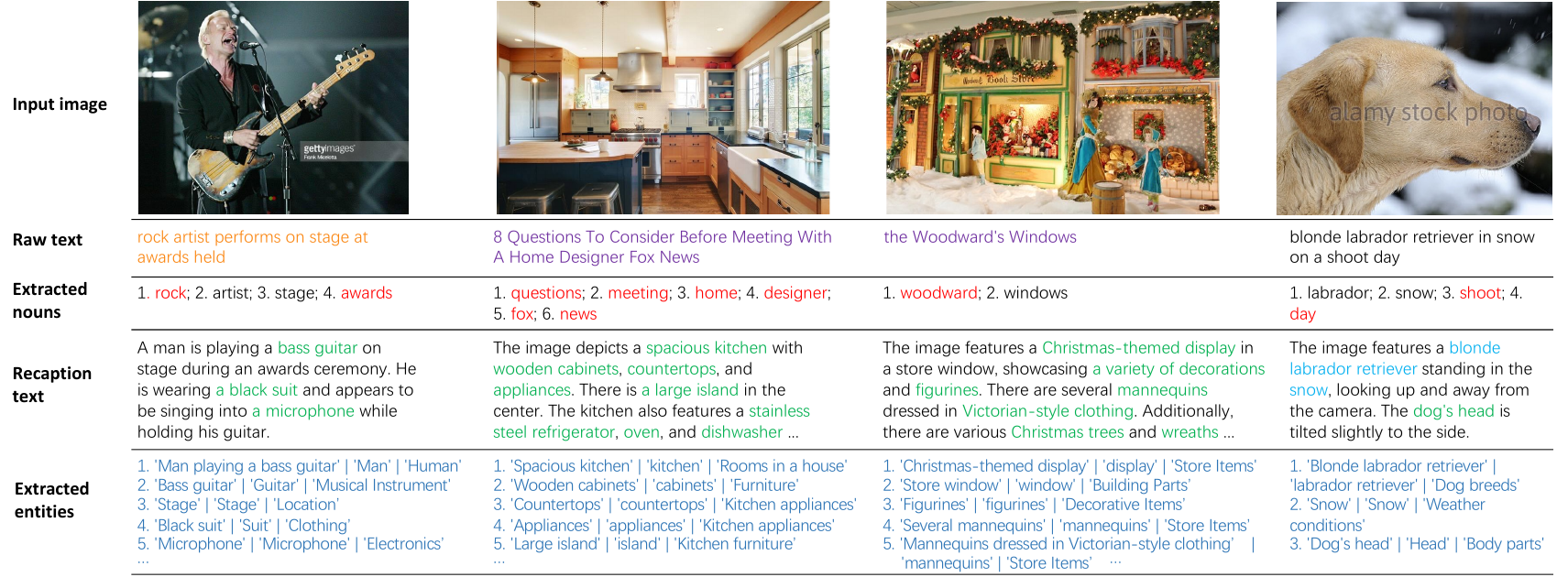} 
\vspace{-1.5em}
\caption{ The illustration of quality issues existing in image-text pair data. \textbf{Row 1}: Existing image-text pair dataset typically suffer from significant \orange{partial annotation} and \deeppurple{image-text misalignment} problems. \textbf{Row 2}: 
Limited by capabilities, traditional NLP parsers~\cite{nltk,spacy} \red{extract nouns do not correspond to actual object in the images}.  \textbf{Row 3}: Our data pipeline provides refined captions with highly detailed image descriptions, \lightblue{preserving effective visual concepts} from the original captions while \forestgreen{supplementing missing concepts}. \textbf{Row 4}: Our data pipeline provides \navyblue{rich, multi-granularity object entity information}.
\label{fig:showcase_image_text}
\vspace{-1em}
}
\end{figure*}

\subsection{Dataset Construction}
\vspace{-.5em}
\label{subsec:dataset_construct}

\paragraph{Auto-annotation data pipeline.}
Leveraging vast, cost-effective image-text pairs for visual concept learning is pivotal in enhancing the generalization capabilities of open-vocabulary object detectors. However, existing image-text pair datasets exhibit significant deficiencies that hinder their utility for OVD, as depicted in Figure~\ref{fig:showcase_image_text}: (1) \textbf{Misalignment}: Internet-sourced image-text pair data frequently contain substantial noise. Even with CLIP~\cite{clip} score-based filtering~\cite{laion400m,laion5b}, many texts still fail to accurately describe the content of images, as shown in the 2nd and 3rd images of Figure~\ref{fig:showcase_image_text}. (2)  \textbf{Partial annotation}: The majority of texts describe only the primary objects in images, resulting in sparse object information, and consequently, hurting learning efficiency of OVD systems, as observed in the 1st image. (3) \textbf{Entity extraction challenge}: Prior works~\cite{detclipv2,vldet,scaleovdet,mdetr} primarily employ conventional NLP parsers, such as NLTK~\cite{nltk,wordnet} or SpaCy~\cite{spacy}, to extract noun concepts from image-text pairs. Their limited capability can result in nouns that poorly align with the images’ content, as illustrated in the second row of Figure~\ref{fig:showcase_image_text}. This mismatch poses further complications for subsequent learning process or pseudo-labeling workflow.

An ideal image-text pair dataset for OVD should encompass accurate and comprehensive descriptions of images, providing information about objects within images across a spectrum of granularity, from detailed to coarse. Motivated by this, we propose the use of a Visual Large Language Model (VLLM)~\cite{llava,instructblip} to develop an automated annotation pipeline, improving the quality of data. VLLMs possess the capability to perceive the contents of images, as well as robust language skills, enabling them to generate precise and detailed captions as well as object descriptions. Specifically, our pipeline comprises the following processes:
\begin{enumerate}
    \item \textbf{Recaptioning with VLLM}: 
    We sample 240k image-text pairs from commonly used datasets~\cite{cc3m,cc12m,yfcc100m} and conducted recaptioning using the InstructBLIP~\cite{instructblip} model. To leverage the information from the original caption, we incorporate it into our prompt design, which is structured as: \textit{``Given a noisy caption of the image: \{raw caption\}, write a detailed clean description of the image."}. This method effectively enhances the quality of the caption texts while maintaining the diversity of noun concepts in the original captions.
    \item \textbf{Entity extraction using GPT-4}: 
    We harness the exceptional language capability of GPT-4~\cite{gpt4} to process entity information in refined captions. Specifically, it is first utilized to filter out non-entity descriptions from the VLLM-generated captions, such as atmospheric or artistic interpretations of the images. Subsequently, it is tasked with extracting object entities present in the captions. Each entity was formatted into a triplet: \textit{\{phrase, category, parent category\}}, representing object descriptions at three distinct levels of granularity.
    \item \textbf{Instruction tuning of VLLM for large-scale annotation}: 
    Considering the substantial costs of the GPT-4 API, its use for large-scale dataset generation is impractical. As a solution, we perform a further instruction tuning phase on a LLaVA~\cite{llava} model, utilizing the improved captions and object entities obtained through prior steps. This finetuned model is then employed to produce captions and entity information for an extensive dataset comprising 200M image-text pairs, sampled from CC15M~\cite{cc3m,cc12m}, YFCC\cite{yfcc100m} and LAION~\cite{laion400m}.
    
    \item \textbf{Auto-labelling for bounding boxes}:
    To automatically derive bounding box annotations for image-text paired data, we apply a pre-trained open-vocabulary object detector (Sec.~\ref{subsec:train strategy}) to assign pseudo bounding box labels given object entities derived from the previous steps. 
    The accuracy of the detector can be greatly improved when provided with accurate candidate object entities from VLLM. 
    Specifically, we utilize the fields \textit{'phrase'} and \textit{'category'} as the textual inputs for the detector and employ a predefined score threshold to filter the resulting bounding boxes. If either of the two fields are matched, we assign the entire entity \textit{\{phrase, category, parent category\}} for that object. After filtering with a predefined confidence threshold, approximately 50M data are sampled for subsequent training, which we refer to as \textbf{\ourdata}. For training the detector, we use the fields of \textit{'phrase'} and \textit{'category'} as the textual label; while for object captioner, we concatenate the three fields -- \textit{'phrase $|$ category $|$ parent category'} -- to serve as the object's ground truth description.
\end{enumerate}

\paragraph{None concept corpus.} Similar to DetCLIP~\cite{detclip}, we develop a noun concept corpus using the information of extracted object entity. This corpus is mainly designed to provide negative concepts for grounding and image-text pair data (Sec.~\ref{subsec:model}). Specifically, we collect the \textit{'category'} field of entities from the 200M recaptioned data. Post frequency analysis, concepts with a total frequency below 10 are omitted. The resulting noun concept corpus of \ourmodel consists of 792k noun concepts, expanding nearly 57 times beyond the 14k concepts built in DetCLIP.

\subsection{Multi-stage Training Scheme}
\label{subsec:train strategy}
Learning to generate diverse object descriptions requires extensive training on large-scale datasets. However, dense prediction tasks such as object detection demand high-resolution inputs to effectively handle scale variance across different objects. This substantially raises the computational cost, posing a challenge to scaling up the training. To mitigate this issue, we develop training strategy based on 'pretraining+finetuning' paradigm to optimize training costs, 
Specifically, it consists of 3 steps:
\input{tables/zero-shot_od}

\begin{enumerate}
    \item \textbf{Training the OV detector (stage 1)}: In the initial phase, we train the OV detector with annotated datasets, \ie, Objects365~\cite{objects365}, V3Det\cite{v3det} and GoldG~\cite{mdetr}. To prepare the model for learning from lower-resolution inputs in later training stages, we apply large-scale jittering augmentation to the training data. Additionally, the model with Swin-L backbone
    developed during this phase is utilized to generate pseudo bounding box for image-text pairs, as described in Sec.~\ref{subsec:dataset_construct}.
    \item \textbf{Pretraining the object captioner (stage 2)}: To enable the object captioner to generate diverse object descriptions, we conduct its pretraining using \ourdata. To boost the efficiency of this training phase, we freeze all parameters of the OV detector, including the backbone, pixel encoder, and object decoder, and adopt a lower input resolution of 320$\times$320. This strategy facilitates the captioner to efficiently acquire the visual concept knowledge from large-scale image-text pairs.
    \item \textbf{Holistic finetuning (stage 3)}: 
    This phase aims to adapt the captioner for high-resolution input while simultaneously improving the OV detector. Specifically, we sample 600k samples from \ourdata with balanced concepts. These samples, along with detection and grounding datasets, are utilized to further fine-tune the model. All parameters are released during this phase to maximize the effectiveness, with the training objective set to  the combination of detection and captioning losses, \ie,  $\mathcal{L} = \mathcal{L}_{det} + \mathcal{L}_{lm}$. The supervision for the captioner solely comes from dataset constructed using our auto-annotation pipeline, whereas all data contribute to the training of the OV detector. Since both the detector and captioner have been pretrained, the model can be efficiently adapted in a few epochs. 
\end{enumerate}

%% file: tables/zero-shot_od.tex
\begin{table*}[th]
\vspace{-4mm}
\centering
\tablestyle{3pt}{1}
\resizebox{\linewidth}{!}{
\begin{tabular}{clllcccccccc}
\toprule
\multirow{2}{*}{}&\multirow{2}{*}{Method} & \multirow{2}{*}{Backbone} & \multirow{2}{*}{Pre-training data}  &  \multicolumn{4}{c}{$\text{LVIS}^{\text{minival}}$} &\multicolumn{4}{c}{$\text{LVIS}^{\text{val}}$} \tabularnewline
& &  &  & $\text{AP}_{\text{all}}$ & $\text{AP}_{\text{r}}$ & $\text{AP}_{\text{c}}$ & $\text{AP}_{\text{f}}$ &$\text{AP}_{\text{all}}$ & $\text{AP}_{\text{r}}$ & $\text{AP}_{\text{c}}$ & $\text{AP}_{\text{f}}$\\
 
\midrule
1 & GLIP~\cite{glip} & Swin-T  & O365,GoldG,Cap4M  & {26.0} & {20.8} & {21.4} & {31.0} &{17.2}&{10.1}&{12.5}&{25.2} \\
2 & GLIPv2~\cite{glipv2} & Swin-T  & O365,GoldG,Cap4M  & {29.0} & {--} & {--} & {--} &{--}&{--}&{--}&{--} \\
3 & CapDet~\cite{capdet} & Swin-T & O365,VG  & {33.8} & {29.6} & {32.8} & {35.5}  &{--}&{--}&{--}&{--} \\
4 & GroundingDINO~\cite{grounddino} & Swin-T  & O365,GoldG,Cap4M  & {27.4} & {18.1} & {23.3} & {32.7}  &{--}&{--}&{--}&{--} \\
5 & OWL-ST~\cite{scaleovdet} & CLIP B/16 & WebLI2B  & {34.4} & {38.3} & {--} & {--}  &{28.6}&{30.3}&{--}&{--} \\
6 & DetCLIP~\cite{detclip} & Swin-T & O365,GoldG,YFCC1M  & {35.9} & {33.2} & {35.7} & {36.4}  &{28.4}&{25.0}&{27.0}&{28.4} \\
7 & DetCLIPv2~\cite{detclipv2} & Swin-T & O365,GoldG,CC15M  & {40.4} & {36.0} & {41.7} & {40.4}  &{32.8}&{31.0}&{31.7}&{34.8} \\
\rowcolor{lightblue} 8 &  \ourmodel  & Swin-T & {O365,V3Det,GoldG,\ourdata}  & \textbf{47.0} & \textbf{45.1} & \textbf{47.7} & \textbf{46.7} &\textbf{38.9}&\textbf{37.2}&\textbf{37.5}&\textbf{41.2} \\
\midrule
9 & GLIP~\cite{glip} & Swin-L  & FourODs,GoldG,Cap24M  & {37.3} & {28.2} & {34.3} & {41.5} &{26.9}&{17.1}&{23.3}&{36.4} \\
10 & GLIPv2~\cite{glipv2} & Swin-H  & \gray{FiveODs},GoldG,CC15M,SBU  & {\gray{50.1}} & {--} & {--} & {--} &{--}&{--}&{--}&{--} \\
11 & GroundingDINO~\cite{grounddino} & Swin-L  & O365,OI,GoldG,Cap4M,\gray{COCO},RefC & {\gray{33.9}} & {22.2} & {30.7} & {38.8} &{--}&{--}&{--}&{--} \\
12 & OWL-ST~\cite{scaleovdet} & CLIP L/14 & WebLI2B  & {40.9} & {41.5} & {--} & {--}  &{35.2}&{36.2}&{--}&{--} \\
13 & DetCLIP~\cite{detclip} & Swin-L & O365,GoldG,YFCC1M  & {38.6} & {36.0} & {38.3} & {39.3} &{28.4}&{25.0}&{27.0}&{31.6} \\
14 & DetCLIPv2~\cite{detclipv2} & Swin-L & O365,GoldG,CC15M  & {44.7} & {43.1} & {46.3} & {43.7} &{36.6}&{33.3}&{36.2}&{38.5} \\
\midrule
\rowcolor{lightblue} 15 & \ourmodel  & Swin-L & {O365,V3Det,GoldG,\ourdata}  & \textbf{48.8} & \textbf{49.9} & \textbf{49.7} & \textbf{47.8}&\textbf{41.4}&\textbf{41.4}&\textbf{40.5}&\textbf{42.3} \\
\bottomrule
\end{tabular}
}
\vspace{-.7em}
\caption{Zero-shot \textit{fixed} AP~\cite{fixap} on LVIS val~\cite{lvis}  and minival~\cite{mdetr}. \gray{Gray} numbers indicate including COCO~\cite{coco} into training, which shares the identical image set with LVIS, thus not representing truly zero-shot. \ourmodel achieves state-of-the-art performance.}
\label{tab:zeroshot_od} \vspace{-1.2em}
\end{table*}

%% file: sec/experiments.tex
\vspace{-.5em}
\section{Experiments}
\vspace{-.5em}
\label{sec: exp}

\paragraph{Training detail.} We train 2 models with Swin-T and Swin-L~\cite{swin_transformer} backbones. The training settings for the object detector primarily follows DetCLIPv2~\cite{detclipv2}. We use 32/64 V100 GPUs to train swin-T/L-based models, respectively. The training epochs for the three phases are 12, 3, and 5, respectively. For the model with the Swin-T backbone, the respective training times for these stages amount to 54, 56, and 35 hours. Refer to Appendix for additional training details.

\subsection{Zero-Shot Open-Vocabulary Object Detection}
\label{subsec:zero-shot ov det}
Following previous works~\cite{glip,glipv2,detclip,detclipv2,scaleovdet}, we evaluate our model's open-vocabulary capability with the zero-shot performance on the 1203-class LVIS~\cite{lvis} dataset. We report performance of \textit{fixed} AP~\cite{fixap} on both val ($\text{LVIS}^{\text{val}}$) and minival~\cite{mdetr} ($\text{LVIS}^{\text{minival}}$) splits. In this experiment, we only use the OV detector component of the model, with class names of the datasets serving as the input.

Table~\ref{tab:zeroshot_od} presents a comparison of our method with existing approaches. \ourmodel significantly outperforms its counterparts, demonstrating superior open-vocabulary object detection capabilities. For instance, on LVIS minival,  our models with Swin-T (row 8) and Swin-L (row 15) backbones achieve an AP of 47.0 and 48.8, respectively, improving upon the previous state-of-the-art method, DetCLIPv2, by 6.6 (row 7) and 4.1 AP (row 14).  Notably, the performance of our Swin-L model on the rare category (49.9 AP) even surpasses that on the base category (47.8 AP in frequent and 49.7 AP in common). This indicates that comprehensive pretraining with high-quality image-text pairs substantially broadens the model's capacity to recognize various visual concepts, leading to significant improved detection capabilities on long-tail distributed data.

\subsection{Evaluation of Object Captioner}
\label{subsec:eval_captioner}

We adopt 2 tasks for evaluating our object captioner, \ie, zero-shot generative object detection and dense captioning.

\paragraph{Zero-shot generative object detection.} 
We conduct zero-shot object-level label generation on COCO~\cite{coco} dataset with the inference process described in Sec.~\ref{subsec:model} and evaluate its detection performance. However, this evaluation poses significant challenges due to two key factors:  (1) the absence of predefined categories for foreground selection results in discrepancies between the detector's proposed foreground regions and dataset's object patterns. (2) the  generation results can be any arbitrary vocabulary, which may not align with the class names specified in the dataset. To mitigate these issues, we introduce several post-processing techniques.
Specifically, we use the 'category' field from the generated labels as the object's class. To address issue (2), during evaluation, we compute similarity between the generated categories and COCO’s class names using the text encoder of the evaluated model, substituting the generated object category with the best-matched COCO category. To resolve issue (1), we further filter out objects with a similarity score below a predefined threshold of 0.7. 

\input{tables/zs_gen_coco}
To compare with existing methods, we adopt the OV COCO setting proposed in OVR-CNN~\cite{ovr-cnn}, where 48 classes from COCO are selected as base classes and 17 as novel classes. The evaluation metric used is the mAP at an IoU of 0.5. Contrary to previous methods,\textit{ we perform zero-shot generative OV detection across all settings without conducting training on the base categories}. Table~\ref{tab:zs_gen_coco} presents the evaluation results. Our generative method can significantly outperform previous discriminative approaches in novel class performance. And our overall AP achieves a level comparable to previous methods without training on base classes. These results demonstrate the potential of generative-based OV detection as a promising paradigm.

\paragraph{Dense captioning.} 
Leveraging visual concept knowledge acquired from extensive image-text pairs, \ourmodel can be easily adapted to generate detailed object descriptions. Following \cite{densecap,TDC}, we evaluate the dense captioning performance on the VG V1.2~\cite{vg} and VG-COCO~\cite{TDC} datasets. To ensure a fair comparison, we finetune our model on the training dataset. Similar to CapDet~\cite{capdet}, during fine-tuning, we convert our OV detector into a class-agnostic foreground extractor, which is achieved by assigning the textual label of all foreground objects to the concept \textit{'object'}. Table~\ref{tab:densecap} compares our method with the existing methods. \ourmodel significantly outperforms existing approaches. \Eg, on VG, our models with Swin-T (row 7) and Swin-L (row 8) backbones surpass the previous best method, GRiT~\cite{grit} (row 6), by 2.9 AP and 4.2 AP, respectively.
\input{tables/densecap}

\subsection{Robustness to Distribution Shift}
\label{subsec:coco-o}

\input{tables/coco-o}
A robust OV object detector should be capable of recognizing a broad spectrum of visual concepts across various domains. The recent vision-language model CLIP~\cite{clip} showcases remarkable generalization to domain shifts in ImageNet variants~\cite{imagenet-a,imagenet-r,imagenet-sketch} through learning from extensive image-text pairs. Similarly, we expect observing comparable phenomena in OV detection. To this end, we use COCO-O~\cite{coco-o} to study our model's robustness to distribution shifts. Table~\ref{tab:coco-o} compares our method with several leading closed-set detectors and the open-set detector GLIP on both COCO and COCO-O. As COCO is not incorporated into our training, \ourmodel's performance trails behind those detectors that are specifically trained on it. However, our model significantly outperforms these detectors on COCO-O. \Eg, our Swin-L model achieves 48.8 AP on COCO-O, which even surpasses its COCO performance (48.5 AP) and attains the best effective robustness score of +27.0. Refer Appendix for qualitative visualizations.

\subsection{Transfer Results with Fine-tuning}
\input{tables/finetune}

Table~\ref{tab:finetune} explores the transferability of \ourmodel by fine-tuning it on downstream datasets, \ie LVIS minival~\cite{mdetr} and ODinW~\cite{glip}. For LVIS, two settings are considered: (1) $\text{LVIS}^\text{mini}_\text{base}$: only base (common and frequent) classes are used for training, as per~\cite{scaleovdet}; and (2) $\text{LVIS}^\text{mini}_\text{all}$: entails training with all categories. 

\ourmodel consistently outperforms its counterparts across all settings. On ODinW13, the Swin-T based \ourmodel (71.1 AP) even surpasses the Swin-L based DetCLIPv2 (70.4 AP). On LVIS, \ourmodel demonstrates exceptional performance, \eg, the Swin-L based model reaches 60.5 AP on both $\text{LVIS}^\text{mini}_\text{base}$ and $\text{LVIS}^\text{mini}_\text{all}$, surpassing OWL-ST+FT~\cite{scaleovdet} (56.2 AP on $\text{LVIS}^\text{mini}_\text{base}$) that pretrains with 2B pseudo-labeled data by a large margin. This indicates the high-quality image-text pairs construted by our auto-annotation pipeline effectively boosts the learning efficiency. Besides, we observe a conclusion parallel to that in ~\cite{scaleovdet}: with strong pretraining, even fine-tuning solely on base categories can substantially enhance performance of rare categories. This is exemplified by the Swin-L model's improvement from 49.8 $\text{AP}_\text{rare}$ in row 15 of Table~\ref{tab:zeroshot_od} to 60.3   $\text{AP}_\text{rare}$ of Table~\ref{tab:finetune}.
\subsection{Ablation Study}
\vspace{-0.5em}
\input{tables/roadmap}
\input{tables/pseudo_label}

\paragraph{\ourmodel's evolution roadmap.} Table~\ref{tab:roadmap} investigates the development roadmap of \ourmodel, from the baseline model to the final version. Our experiments utilize a model with a Swin-T backbone. For the OV detector, we evaluate the AP on LVIS minival (Sec.~\ref{subsec:zero-shot ov det}) and COCO-O (Sec.~\ref{subsec:coco-o}), and for the captioner, we report the fine-tuned performance on VG (Sec.~\ref{subsec:eval_captioner}). Our baseline (row 1) model is our OV detector (as described in Sec.~\ref{subsec:model}) without the object captioner and is trained solely on the Objects365~\cite{objects365}. This model exhibits limited capabilities, achieving a modest 30.8 AP on LVIS. Subsequently, we introduce a series of effective designs: 
\textbf{(1) Incorporating more human-annotated data} (rows 2 and 3), \ie, GoldG~\cite{mdetr} and V3Det~\cite{v3det}, significantly boosts the LVIS AP to 42.5. \textbf{(2) Introducing image-text pair data}, \ie, 600k samples from GranuCap50M (also the training data used in our stage 3 training, see Sec.~\ref{subsec:train strategy}), effectively further improve the LVIS AP to 45.3. More importantly, it significantly improve the model's domain generalization, bring COCO-O AP from 30.7 in row3 to 36.4 in row 4. \textbf{(3) } Row 5 further \textbf{integrates the object captioner}, yet without the stage 2 pretraining. It boosts the LVIS AP to 46.6, despite no new data is introduced. This improvement reveals the learning of captioner benefits OV detection -- learning to generate diverse labels for objects encourages the object decoder to extract more discriminative object features. \textbf{(4)  Integrating stage 2 captioner pretraining} efficiently acquires broad visual concept knowledge from the massive image-text pairs of GranuCap50M. This design significantly enhances the generative capability of the captioner, boosting the VG AP from 17.1 of row 5 to 18.4 of row 6. Furthermore, it modestly improve the OV detection performance from 46.6 AP to 47.0 AP on LVIS.

\paragraph{Pseudo-labeling for Image-text Pairs.} Table~\ref{tab:pseudo_label} investigates two critical factors in utilizing pseudo-labeled image-text pairs: the filtering threshold and the data volume. We experiment with Swin-T model for stage 1 training, with pseudo-label data incorporated. A filtering threshold of 0.2 achieved the best results, and increasing the volume of data continuously improved the performance of the OV detection. Though incorporating 1200k data achieving better results, we opt for 600k data for stage 3 training for efficiency consideration. Notably, when assisted with the captioner's learning in generative tasks, the effectiveness of 600k data samples (row 5 of Table~\ref{tab:roadmap}, 46.6 AP) surpasses the result of 1200k samples without the captioner (46.1 AP).

\subsection{Visualization}
Figure \ref{fig:det_gen_results} provides visualization results for both OV detection and object label generation of \ourmodel. Our model demonstrates superior visual understanding capabilities, capable of detecting or generating a broad range of visual concepts. Refer to Appendix for more visualization results.

%% file: tables/zs_gen_coco.tex
\begin{table}[t!]
\begin{center}
\resizebox{\linewidth}{!}{
\begin{tabular}{lcccc}
\toprule 
{Method} & Generative & \makecell{Novel \\ $\text{AP}_\text{50}$} & \makecell{Base \\ $\text{AP}_\text{50}$} & \makecell{Overall  \\ $\text{AP}_\text{50}$} \\
\midrule
OVR-CNN~\cite{ovr-cnn} & \xmark & 22.8 & \gray{46.0} & \gray{39.9} \\
Detic~\cite{detic} & \xmark & 27.8 & \gray{47.1} & \gray{42.0} \\
RegionCLIP~\cite{regionclip} & \xmark & 26.8 & \gray{54.8} & \gray{47.5} \\
ViLD~\cite{vild} & \xmark & 27.6 & \gray{59.5} & \gray{51.3} \\
VLDet~\cite{vldet} & \xmark & 32.0 & \gray{50.6} & \gray{45.8} \\
DetPro~\cite{detpro} & \xmark & 43.3 & \gray{61.9} & \gray{55.7} \\
\midrule
\rowcolor{lightblue}\ourmodel (Swin-T) & \cmark & 54.7 & 42.8 & 46.9 \\
\rowcolor{lightblue}\ourmodel (Swin-L) &\cmark & \textbf{57.3} & 44.2 & 49.3\\
\bottomrule
\end{tabular}
}
\vspace{-.7em}
\caption{Zero-shot generative object detection on COCO~\cite{coco}. \gray{Gray} numbers indicate training on COCO's base classes. Our method significantly outperforms previous OV detectors in novel categories in a generative manner.}
\label{tab:zs_gen_coco}
\vspace{-2em}
\end{center}
\end{table}

%% file: tables/densecap.tex
\begin{table}[t]
\begin{center}
\begin{tabular}{clcc}
\toprule 
& {Method} & \makecell{VG V1.2 \\ AP} & \makecell{VG-COCO \\ AP}\\
\midrule
1 & JIVC~\cite{densecap} & 10.0 & 7.9 \\
2 & COCG~\cite{imgG} & 10.4 & 8.9 \\
3 & CAG-Net~\cite{CAGNet} & 10.5 & {--} \\
4 & TDC+ROCSU~\cite{TDC} & 11.9 & 11.6 \\
5 & CapDet~\cite{capdet} & 15.4 & 14.0 \\
6 & GRiT~\cite{grit} & 15.5 & {--}  \\
\midrule
\rowcolor{lightblue}7 & {\ourmodel (Swin-T)}& {18.4} & {17.7} \tabularnewline
\rowcolor{lightblue}8 & {\ourmodel (Swin-L)}& \textbf{19.7} & \textbf{18.9} \tabularnewline
\bottomrule
\end{tabular}
\vspace{-.7em}
\caption{Dense captioning on VG V1.2~\cite{vg} and VG-COCO~\cite{TDC}.}
\label{tab:densecap}
\vspace{-3em}
\end{center}
\end{table}

%% file: tables/coco-o.tex
\begin{table}[t!]
\begin{center}
\resizebox{\linewidth}{!}{
\begin{tabular}{lcccc}
\toprule 
{Method} & Backbone & \makecell{COCO \\ AP} & \makecell{COCO-O \\ AP} & \makecell{Effective \\ Robustness} \\
\midrule
GLIP~\cite{glip} & Swin-T & 46.1 & 29 & \darkgreen{+8.0} \\
\rowcolor{lightblue}{\ourmodel} & Swin-T  & \textbf{47.2}  & \textbf{38.5}  & \textbf{\darkgreen{+17.3}} \\
\midrule
DINO~\cite{dino} & Swin-L & \gray{58.5} &  42.1 & \darkgreen{+15.8} \\
DyHead~\cite{dyhead} & Swin-L & \gray{56.2} & 35.3 & \darkgreen{+10.0} \\
GLIP~\cite{glip} & Swin-L & 51.4 & 48 & \darkgreen{+24.9} \\
GRiT~\cite{grit} & ViT-H & \gray{60.4} & 42.9 & \darkgreen{+15.7} \\
\rowcolor{lightblue}{\ourmodel} & Swin-L  & {48.5}  & \textbf{48.8}  & \textbf{\darkgreen{+27.0}} \\
\bottomrule
\end{tabular}
}
\vspace{-.7em}
\caption{Distribution shift performance on  COCO-O~\cite{coco-o}. \gray{Gray} numbers indicate include COCO~\cite{coco} data into training.}
\label{tab:coco-o}
\vspace{-2em}
\end{center}
\end{table}

%% file: tables/finetune.tex
\begin{table}[t]

\centering
\resizebox{\linewidth}{!}{
\begin{tabular}{lcccccc}
\toprule
\multirow{2}{*}{Method} & \multirow{2}{*}{Backbone}& \multicolumn{2}{c}{$\text{LVIS}^\text{mini}_\text{base}$} & \multicolumn{2}{c}{$\text{LVIS}^\text{mini}_\text{all}$} & ODinW13 \tabularnewline
 & & {$\text{AP}_\text{all}$} & {$\text{AP}_\text{rare}$} & {$\text{AP}_\text{all}$} & {$\text{AP}_\text{rare}$} & AP \\
\midrule
GLIP \cite{glip} & Swin-T & {--} & {--} & {--} & {--} & 64.9 \\
GLIPv2 \cite{glipv2} & Swin-T & {--} & {--} & {50.6}   & {--} & 66.5 \\
DetCLIPv2 \cite{detclipv2} & Swin-T & {--} & {--} & {50.7} & {44.3}  & 68.0 \\
OWL-ST+FT\cite{scaleovdet} & CLIP B/16 & {48.7} & {42.1} & {--} & {--}  & {--} \\
\rowcolor{lightblue}  \ourmodel  & Swin-T & \textbf{54.3} & \textbf{53.7} & \textbf{56.5} & \textbf{55.1} & \textbf{71.1} \\
\midrule
GLIP \cite{glip} & Swin-L & {--} & {--} & {--} & {--} &  68.9 \\
GLIPv2 \cite{glipv2} & Swin-H & {--} & {--} & {59.8} & {--}   &  {70.4} \\
DetCLIPv2\cite{detclipv2} & Swin-L & {--} & {--} & {60.1} & {58.3}   &  {70.4}  \\
OWL-ST+FT\cite{scaleovdet} & CLIP L/14 & {56.2} & {52.3} & {--} & {--}  & {--} \\
\rowcolor{lightblue}  \ourmodel & Swin-L  & \textbf{60.5} & \textbf{60.3} & \textbf{60.5} & \textbf{60.7} & \textbf{72.1} \\
\bottomrule
\end{tabular}
}
\vspace{-.7em}
\caption{Fine-tuning performance. \textit{Fixed} AP \cite{fixap} on LVIS minival5k \cite{mdetr} and average AP on ODinW13 \cite{glip} are reported.}
\label{tab:finetune} \vspace{-1.3em}
\end{table}

%% file: tables/roadmap.tex
\begin{table}[t!]
\begin{center}
\resizebox{\linewidth}{!}{
\begin{tabular}{clcccc}
\toprule 
  & & \multicolumn{2}{c}{$\text{LVIS}^{\text{mini}}$}  & COCO-O & VG V1.2 \\
  & & $\text{AP}_{\text{all}}$ & $\text{AP}_{\text{rare}}$  &  AP & AP \\
\midrule
1 & Baseline & 30.8  & 28.7 & 24.1 & {--} \\
2 & +GoldG & 41.4  & 37.5 & 32.5 & {--} \\
3 & +V3Det   &  42.5 & 39.4 & 30.7 & {--} \\
4 &+GranuCap600k & 45.3 & 42.2 & 36.4 & {--} \\
5 &+Captioner & 46.6 & 44.5 & 38.0 & {17.1} \\
6 &+Stage2 pretraining & \textbf{47.0} & \textbf{45.1} & \textbf{38.5}&\textbf{18.4} \\

\bottomrule
\end{tabular}
}
\vspace{-.7em}
\caption{The evolution roadmap of \ourmodel. Each row introduces new changes building upon the results of the preceding row.}
\label{tab:roadmap}
\vspace{-2em}
\end{center}
\end{table}

%% file: tables/pseudo_label.tex
\begin{table}[t!]
\begin{center}
\resizebox{0.8\linewidth}{!}{
\begin{tabular}{ccccc}
\toprule 
& \multirow{2}{*}{Threshold} & \multirow{2}{*}{$\#$Samples} & \multicolumn{2}{c}{$\text{LVIS}^{\text{mini}}$} \\
& & & $\text{AP}_{\text{all}}$ & $\text{AP}_{\text{rare}}$  \\
\midrule
1 & 0.15 & 600k & 45.0 & \textbf{43.2}  \\
2 & 0.2 & 600k & \textbf{45.3} & 42.2  \\
3 & 0.25 & 600k & 44.8 & 42.3  \\
4 & 0.3 & 600k & 44.9 & 41.0  \\
\midrule
5 & 0.2 & 300k & 44.0 & 40.1  \\
6 & 0.2 & 600k & 45.3 & 42.2  \\
7 & 0.2 & 1200k & \textbf{46.1} & \textbf{44.2}  \\
\bottomrule
\end{tabular}
}
\vspace{-.7em}
\caption{Impact of filtering threshold and data volume of pseudo-labeled image-text pairs.}
\label{tab:pseudo_label}
\vspace{-2em}
\end{center}
\end{table}

%% file: sec/conclusion.tex
\vspace{-0.5em}
\section{Limitation and Conclusion}
\vspace{-0.5em}
\label{sec: conclusion}
\paragraph{Limitation.} The evaluation of DetCLIPv3's generative capability remains incomplete, as existing benchmarks fall short in effectively evaluating generative detection results. Moreover, the detection process in \ourmodel currently does not support control via instructions. Moving forward, an important research direction will be to develop comprehensive metrics for evaluating generative open-vocabulary detectors and to integrate large language models (LLMs) for instruction-controled open-vocabulary detection.

\paragraph{Conclusion.} In this paper, we present \ourmodel, an innovative OV detector that is capable of localizing objects based on category names, as well as generating object label that is hierarchical and multi-granular. Such enhanced visual capability enables more comprehensive fine-grained visual understanding, which expands the application scenarios for OVD model. We hope our method provides insights for future development of visual cognition systems.

%% file: sec/appendix.tex
\appendix

\section{Additional Implementation Details}
\paragraph{Training.}
The training of \ourmodel involves data from various sources. Table~\ref{app:tab:dataset_info} summarizes detailed data information used in different training phases. Since the training process varies for different data type, (\eg, the object captioner accepts only image-text pair data as input), we design each iteration's global batch to contain only one type of data.

For the training of the the open-vocabulary detector, following previous DetCLIP works~\cite{detclip,detclipv2}, we initialize the text encoder with the parameters of FILIP's~\cite{filip} language model, and reduce its learning rate by 0.1 during training to preserve the knowledge obtained via FILIP’s pre-training. To improve training efficiency, we set the maximum text token length for the text encoder to 16. 

For the training of the object captioner, we initialize the captioner with the pre-trained weights of Qformer~\cite{blip2}, whereas the deformable~\cite{deformdetr} cross-attention layers are randomly initialized. To preserve the knowledge acquired during Qformer~\cite{blip2} pretraining, the object captioner utilizes the same BERT~\cite{bert} tokenizer for processing text input, different to the text encoder which employs the CLIP~\cite{clip} tokenizer. The maximum text token length for the object captioner is set to 32.

In each training stage, to conserve GPU memory, automatic mixed-precision~\cite{amp} and gradient checkpointing~\cite{grad_checkpoint} are employed. Table~\ref{app:tab:training_detail} summarizes the detailed training settings for each training stage.

\input{tables/appendix/dataset_info}
\input{tables/appendix/training}

\input{tables/appendix/finetune_detail}
\paragraph{Inference.} Inference process of \ourmodel's OV detector follows DINO~\cite{dino}, where the results for each image are derived from the predictions of 300 object queries with highest confidence scores. For the \textit{fixed} AP~\cite{fixap} evaluation on the LVIS~\cite{lvis} dataset, it is required that each category within the entire validation set has at least 10,000 predictions. To ensure an sufficient number of predictions per image, we adopt an inference process similar to that of GLIP~\cite{glip}. Specifically, during inference for each data sample, the 1203 categories are split into 31 chunks, with a chunk size of 40 categories. We conduct inference separately for each chunk and retain the top 300 predictions based on their confidence scores.

For the inference process of \ourmodel's object captioner, as described in the main paper, for each image, we utilize the most frequent 15k concepts from our developed noun concept corpus as text queries to extract top 100 foreground regions with highest similarity. After the generation of descriptive labels for these regions by the object captioner, their confidence scores are re-calibrated using the OV detector. A class-agnostic non-maximum suppression (NMS) operation is then performed for regions with re-calibrated scores higher than 0.05, the results of which are output as predictions. We set beam search's beam size equal to 1 for inference of object captioner.

\paragraph{Finetuning.} We fine-tune \ourmodel on 2 datasets, \ie, LVIS \cite{lvis} and ODinW13 \cite{glip}.   Table~\ref{app:tab:finetune_lvis_details} and~\ref{app:tab:finetune_odinw_details} summarize the detailed fine-tuning settings for LVIS and ODinW13, respectively.
For LVIS, when fine-tuning with base categories, we exclude novel categories while sampling negative concepts.
For ODinW13, similar to DetCLIPv2\cite{detclipv2}, we employ an auto-decay learning rate schedule. Specifically, when performance reaches a plateau and persists for a tolerance period $t_1$, we reduce the learning rate by a factor of 0.1. If there is no improvement in performance for a tolerance period $t_2$, we then terminate the training process.

\begin{figure*}
\includegraphics[width=1.0\textwidth]{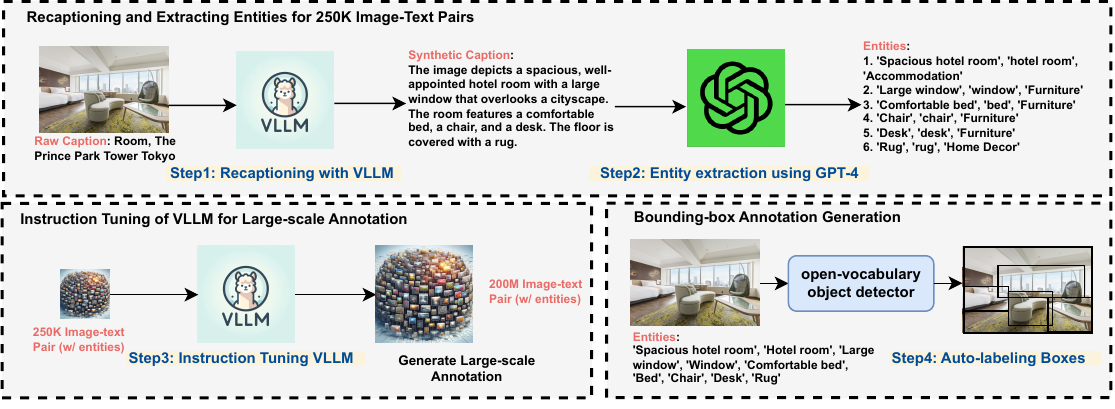} 
\vspace{-1.5em}
\caption{ The illustration for \ourmodel's auto-annotation data pipeline. It initially utilizes a VLLM to recaption 240K image-text pairs, followed by the use of GPT-4 to extract object entities, formatted as \textit{\{phrase, category, parent category\}}. Subsequently, these image-text pairs are used for instruction tuning of a VLLM, with the trained model providing annotations for a larger-scale of 200M image-text pairs. Finally, an OV detector is employed to provide pseudo-label bounding boxes for these data, and after applying a confidence score filtering, 50M data are sampled to form \ourdata.}
\label{fig:data_pipeline}
\vspace{-.5em}
\end{figure*}

\section{Additional Data Pipeline Details}
Figure~\ref{fig:data_pipeline} illustrates an overview of \ourmodel's auto-annotation data pipeline.

\paragraph{Prompts.} Here we provide the prompts used in each step, including those for the VLLMs as well as for GPT-4.
\input{tables/appendix/prompts}

\paragraph{Visualizations.} Figure~\ref{fig:pipeline_cases-1} and~\ref{fig:pipeline_cases-2} depict refined caption and extracted entity information obtained via our proposed data pipeline. Additionally, Figure~\ref{fig:pdet_label} displays the bounding box pseudo-labels generated by our Swin-L-based model after stage-1 training.

\renewcommand{\thefigure}{2-a}
\begin{figure*}
\vspace{-2em}
\includegraphics[width=1.0\textwidth]{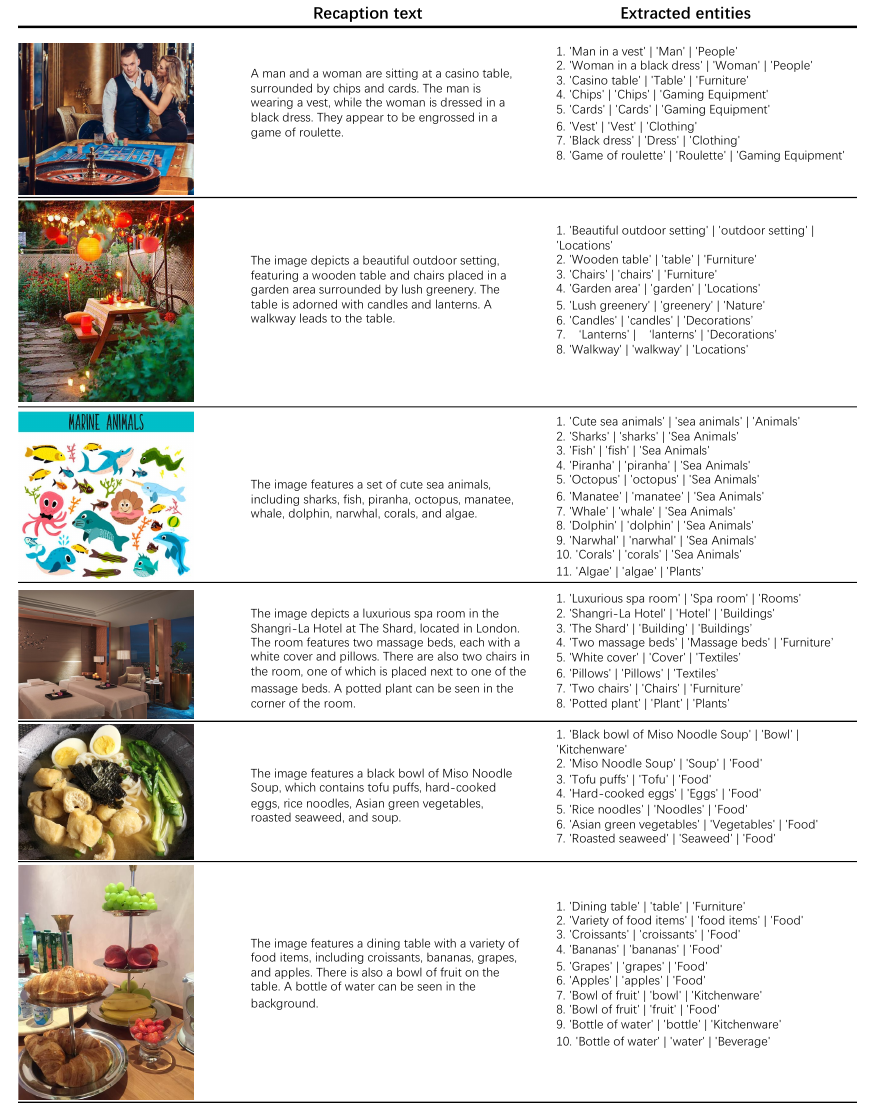} 
\vspace{-1.5em}
\caption{Examples of refined captions and extracted object entities yield by \ourmodel's data pipeline.}
\label{fig:pipeline_cases-1}
\end{figure*}

\renewcommand{\thefigure}{2-b}
\begin{figure*}
\vspace{-2em}
\includegraphics[width=1.0\textwidth]{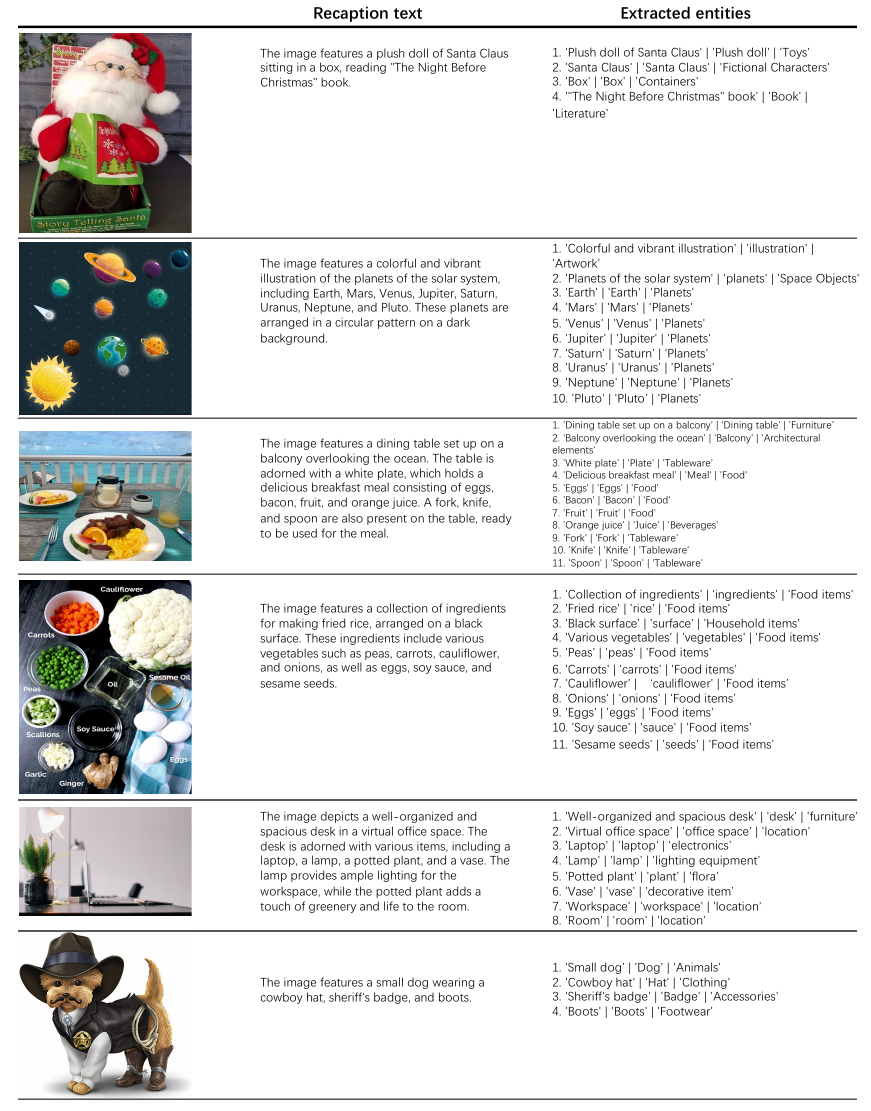} 
\vspace{-1.5em}
\caption{Examples of refined captions and extracted object object entities yield by \ourmodel's data pipeline.}
\label{fig:pipeline_cases-2}
\end{figure*}

\setcounter{figure}{2}
\renewcommand{\thefigure}{\arabic{figure}}

\begin{figure*}
\vspace{-2.5em}
\includegraphics[width=1.0\textwidth]{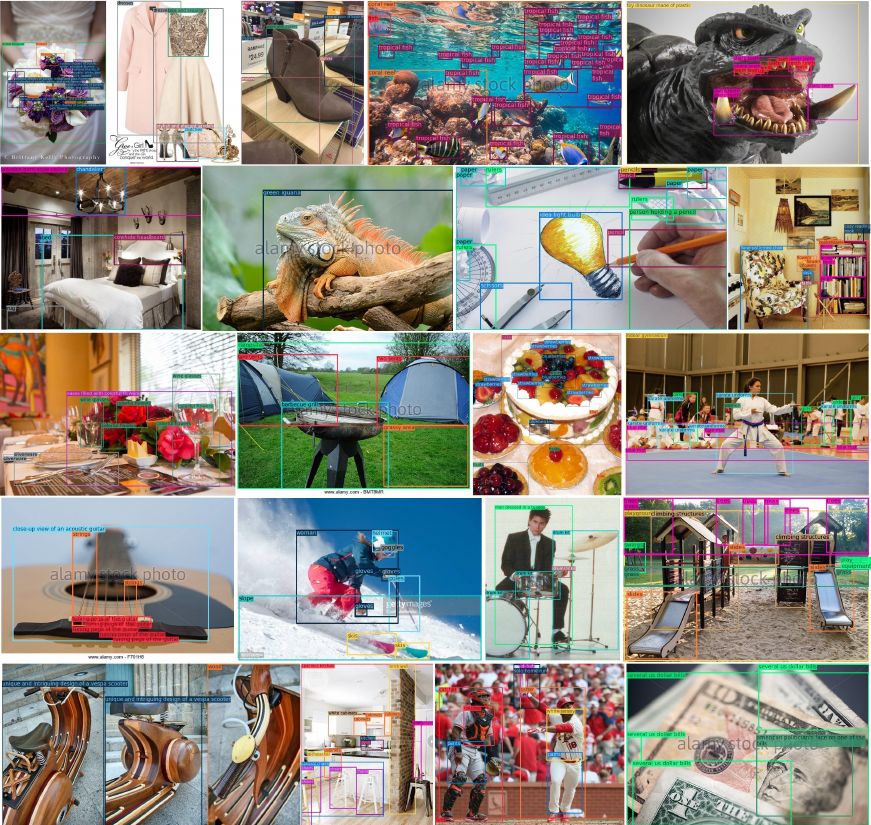} 
\vspace{-2em}
\caption{Examples of bounding box pseudo-labels generated by \ourmodel's Swin-L model after stage-1 training.}
\label{fig:pdet_label}

\end{figure*}

\input{tables/appendix/lvis}
\input{tables/appendix/coco-o}
\input{tables/appendix/odinw}
\section{More Qualitative Results}
Figure~\ref{fig:obj_caption-1.pdf},~\ref{fig:obj_caption-2.pdf} and~\ref{fig:obj_caption-3.pdf} present additional qualitative results showcasing multi-granular object labels generated by \ourmodel's object captioner. In the absence of candidate categories, \ourmodel's object captioner generates dense, fine-grained, multi-granular object labels, thus facilitating a more comprehensive image understanding.

\renewcommand{\thefigure}{4-a}
\begin{figure*}
\vspace{-2.5em}
\includegraphics[width=1.0\textwidth]{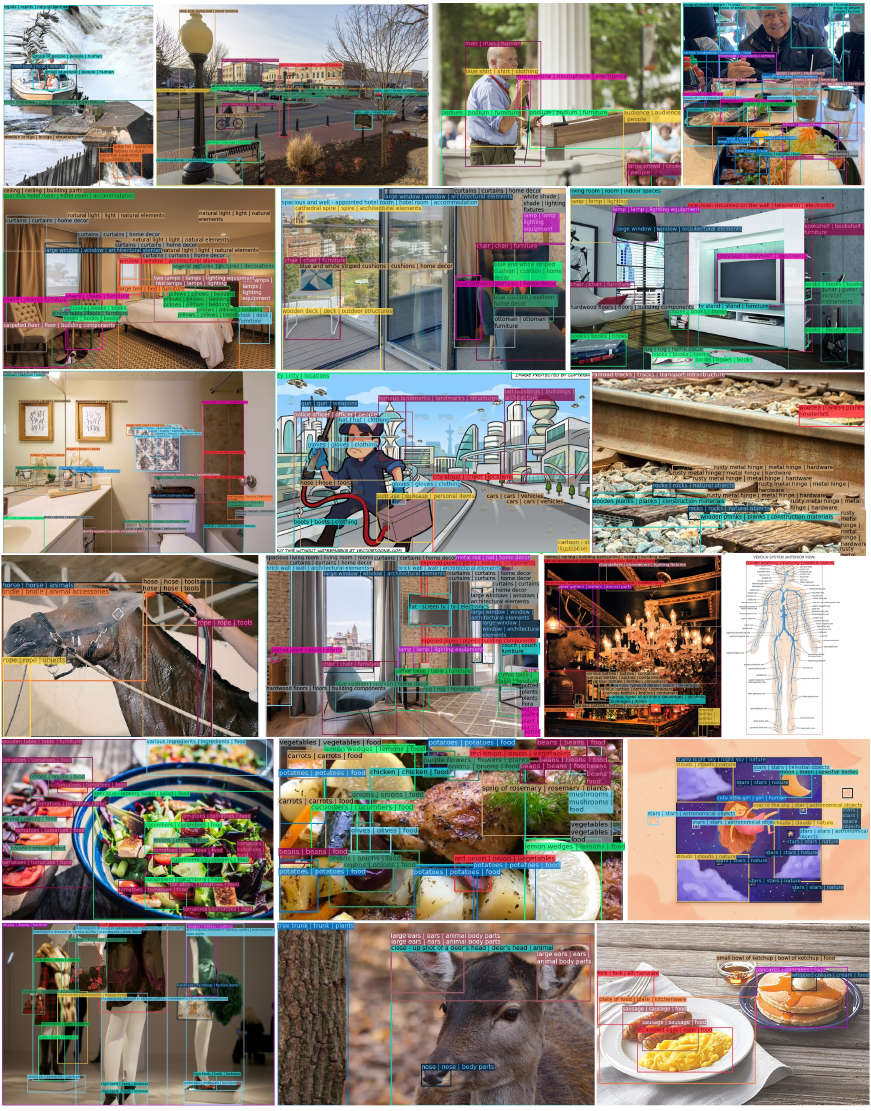} 
\vspace{-2em}
\caption{Qualitative results of multi-granular object labels generated by \ourmodel's object captioner. In the absence of candidate categories, \ourmodel's object captioner generates dense, fine-grained, multi-granular object labels, thus facilitating a more comprehensive image understanding. }
\label{fig:obj_caption-1.pdf}
\end{figure*}

\renewcommand{\thefigure}{4-b}
\begin{figure*}
\vspace{-2.5em}
\includegraphics[width=1.0\textwidth]{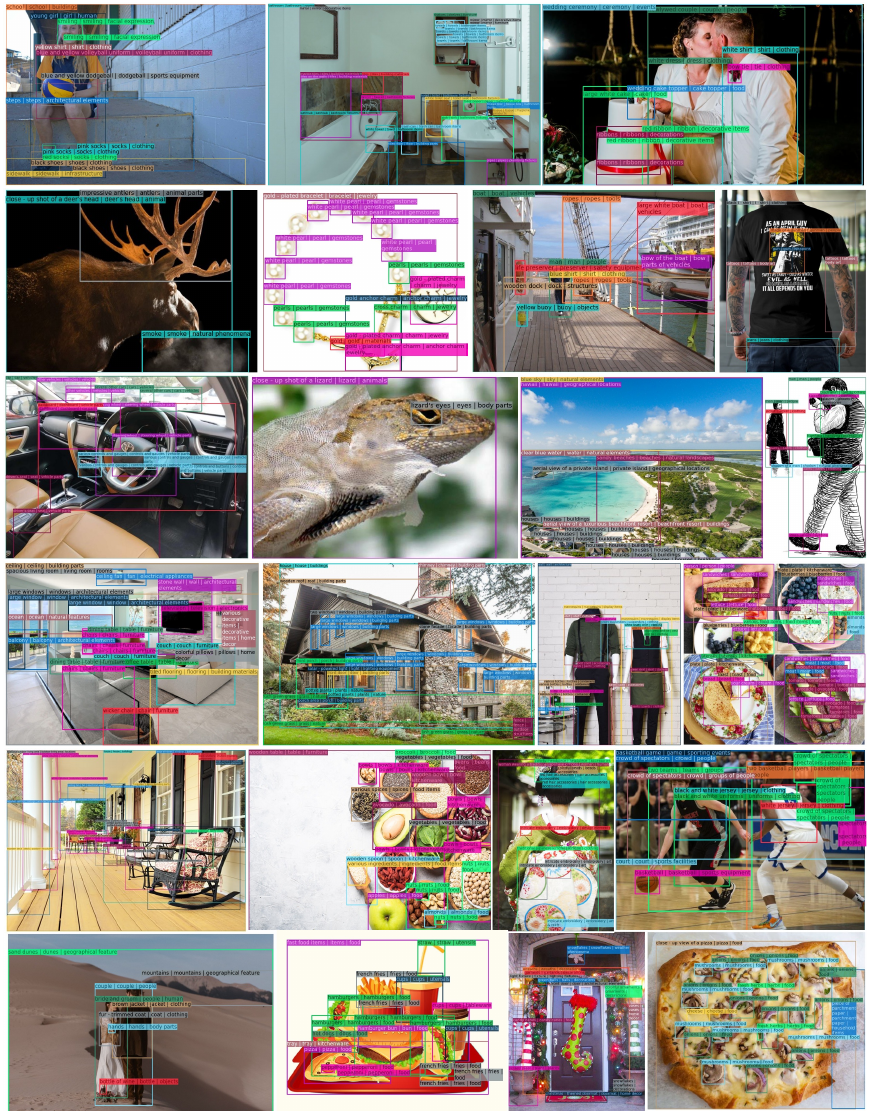} 
\vspace{-2em}
\caption{Qualitative results of multi-granular object labels generated by \ourmodel's object captioner. In the absence of candidate categories, \ourmodel's object captioner generates dense, fine-grained, multi-granular object labels, thus facilitating a more comprehensive image understanding. }
\label{fig:obj_caption-2.pdf}
\end{figure*}

\renewcommand{\thefigure}{4-c}
\begin{figure*}
\vspace{-2.5em}
\includegraphics[width=1.0\textwidth]{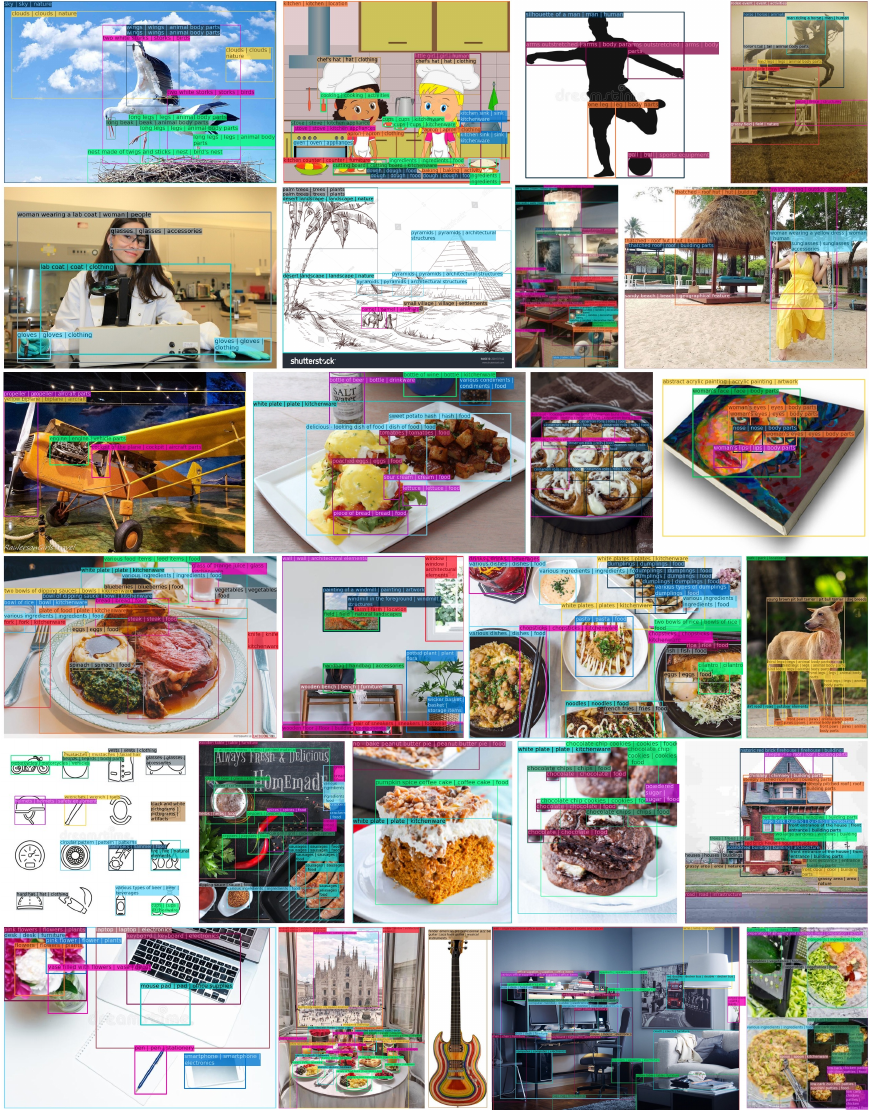} 
\vspace{-2em}
\caption{Qualitative results of multi-granular object labels generated by \ourmodel's object captioner. In the absence of candidate categories, \ourmodel's object captioner generates dense, fine-grained, multi-granular object labels, thus facilitating a more comprehensive image understanding. }
\label{fig:obj_caption-3.pdf}
\end{figure*}

\setcounter{figure}{4}
\renewcommand{\thefigure}{\arabic{figure}}

\section{More Experimental Results}
\begin{figure*}
\includegraphics[width=1.0\textwidth]{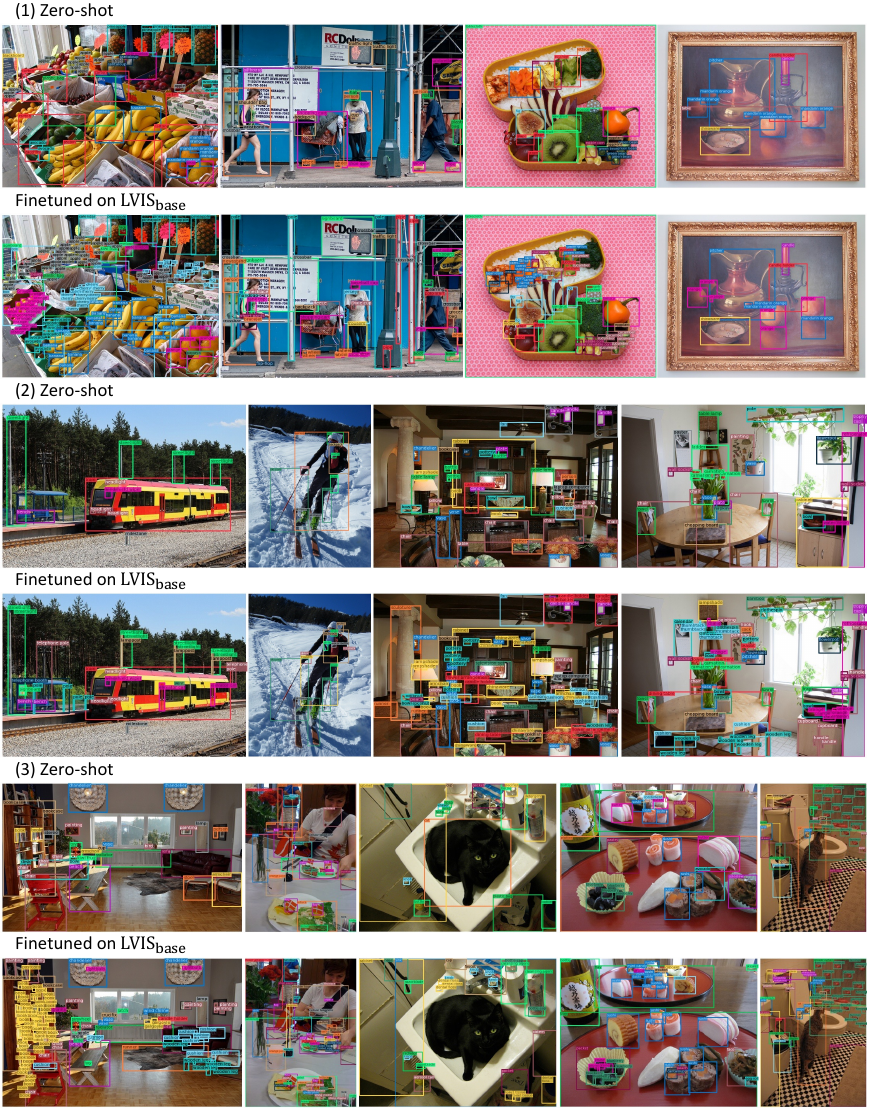} 
\vspace{-1.5em}
\caption{Visualization of detection results on LVIS~\cite{lvis}. In each group, the first row represents the zero-shot results, while the second row indicates the results after fine-tuning on the LVIS base categories.}
\label{fig:lvis_results}
\end{figure*}

\paragraph{More results on LVIS.} 
To comprehensively evaluate the performance of \ourmodel, Table~\ref{app:tab:lvis} provides the standard Average Precision (AP) on LVIS, comparing it with the state-of-the-art method OWL-ST~\cite{scaleovdet}, which is pretrained on 2 billion image-text pairs. Specifically, we assess two settings on the LVIS minival~\cite{mdetr} and validation~\cite{lvis} datasets: the zero-shot performance and the performance after fine-tuning on LVIS base categories. Despite being pretrained with only 50M image-text pairs, \ourmodel markedly outperforms OWL-ST, \eg, \ourmodel surpasses OWL-ST's counterparts by over 5 AP across all settings, demonstrating the superior learning efficiency of our method. Figure~\ref{fig:lvis_results} provides the detection results on both zero-shot and $\text{LVIS}_\text{base}$ fine-tuning settings.

\begin{figure*}
\vspace{-2em}

\renewcommand{\thefigure}{6-a}
\includegraphics[width=1.0\textwidth]{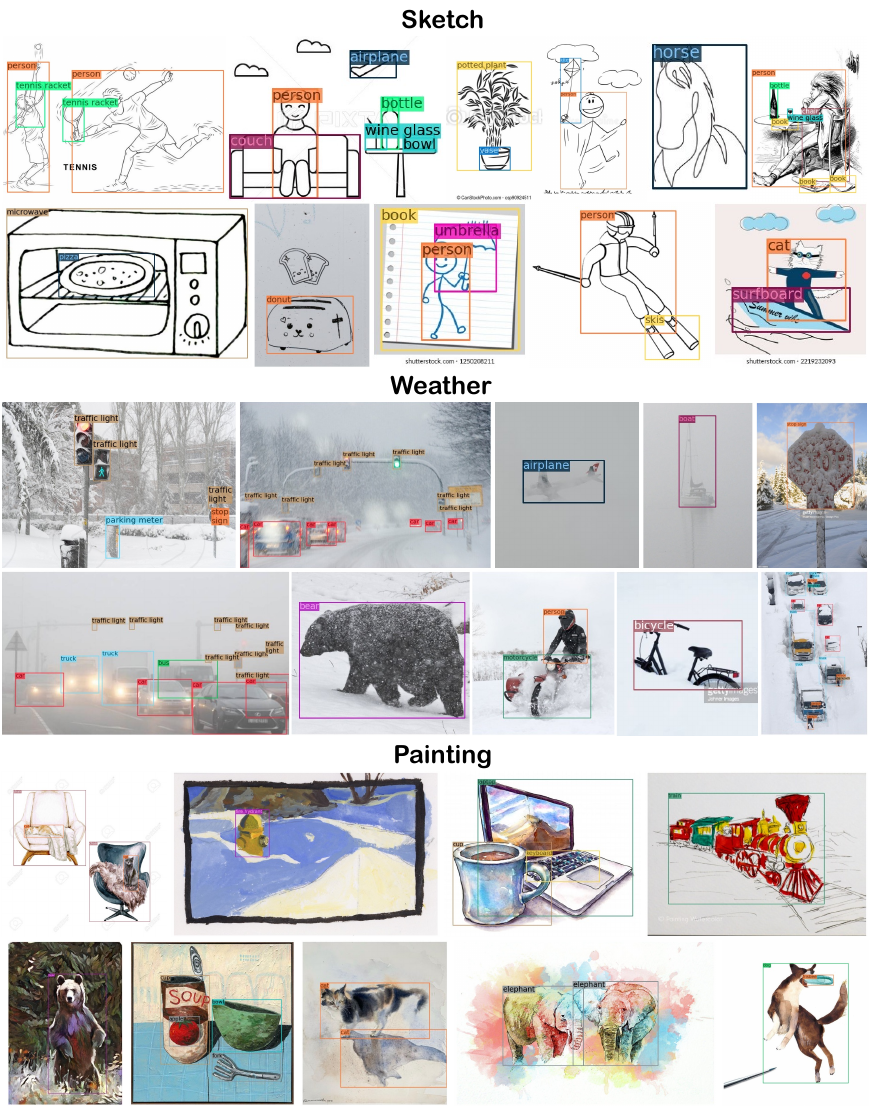} 
\vspace{-1.5em}
\caption{Zero-shot detection results on COCO-O\cite{coco-o} dataset. \ourmodel exhibits a robust domain generalization capability.}
\label{fig:coco-o-1}
\end{figure*}

\paragraph{Detailed performance on COCO-O.} Table~\ref{app:tab:coco-o_full_results} reports the detailed zero-shot AP performance on COCO-O~\cite{coco-o}'s 6 domains, \ie, sketch, weather, cartoon, painting, tattoo and handmake. Figure~\ref{fig:coco-o-1} and~\ref{fig:coco-o-2} visualizes the detection results, demonstrating \ourmodel's robust domain generalization capability.

\renewcommand{\thefigure}{6-b}
\begin{figure*}
\vspace{-2em}
\includegraphics[width=1.0\textwidth]{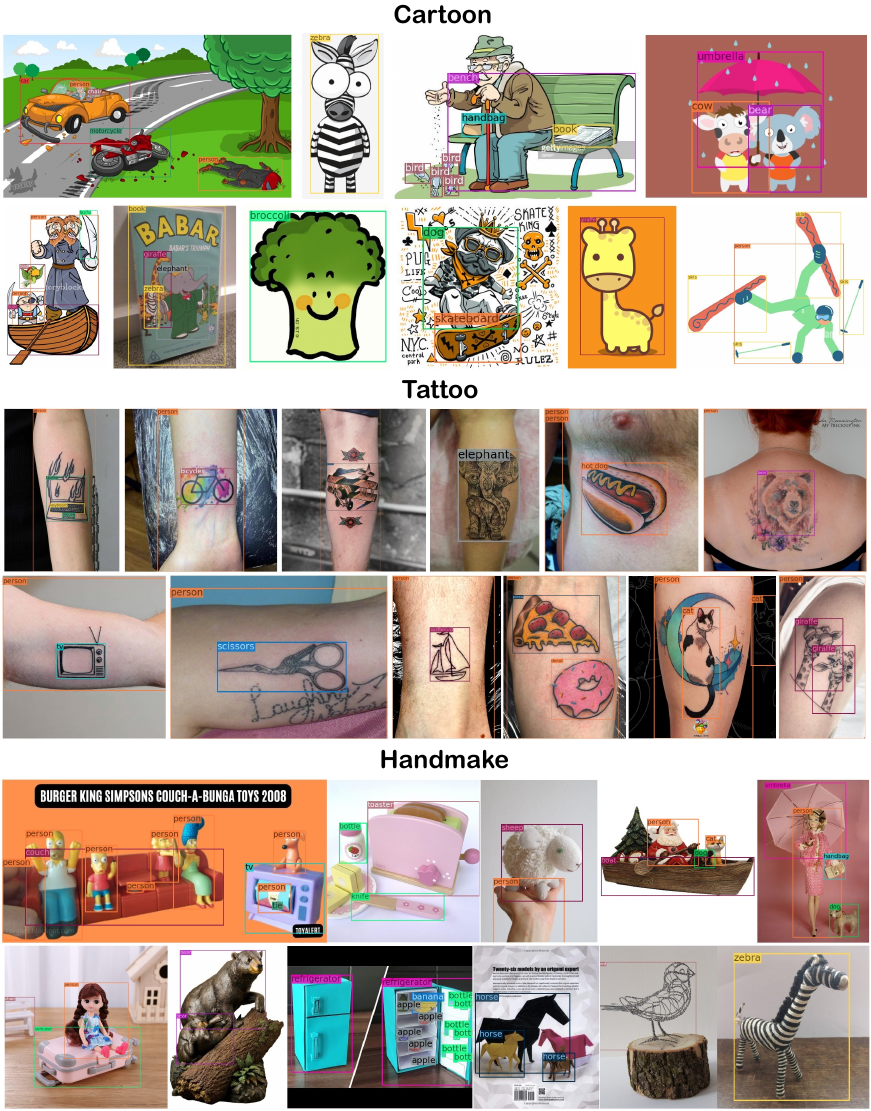} 
\vspace{-1.5em}
\caption{Zero-shot detection results on COCO-O\cite{coco-o} dataset. \ourmodel exhibits a robust domain generalization capability.}
\label{fig:coco-o-2}
\end{figure*}

\paragraph{Detailed performance on ODinW13.} Table~\ref{app:tab:odinw-full-results} reports the detailed fine-tuned performance on ODinW13~\cite{glip} dataset.

\input{tables/appendix/inference_speed}
\paragraph{Inference speed.} Table~\ref{app:tab:inference speed} reports the inference speed of \ourmodel as well as a comparison with previous methods.

%% file: tables/appendix/dataset_info.tex
\begin{table}[h]
\begin{center}
\resizebox{\linewidth}{!}{
\begin{tabular}{c|ccc}
\toprule
 \makecell{Training Stage} & Dataset  & Total Volume  \\
\midrule
Stage 1 & \makecell{O365~\cite{objects365}(0.66M), GoldG~\cite{glip}(0.77M),\\ V3Det~\cite{v3det}(0.18M)} & 1.61M \\
\midrule
Stage 2 & \ourdata & 50M \\
\midrule
Stage 3 & \makecell{O365~\cite{objects365}(0.66M), GoldG~\cite{glip}(0.77M),\\ V3Det~\cite{v3det}(0.18M), \ourdatasub (0.6M)} & 2.21M \\
\bottomrule
\end{tabular}
}
\end{center}
\caption{Dataset information for each training stage of \ourmodel. O365 refers to the Objects365 v2, from which we sample 0.66M data with balanced class for training, similar to previous DetCLIP v1/v2~\cite{detclip,detclipv2} works. GranuCap50M is developed with our proposed auto-annotation pipeline, using 50M image-text pairs sampled from a collection of CC3M~\cite{cc3m}, CC12M~\cite{cc12m}, YFCC100M~\cite{yfcc100m} and LAION400M~\cite{laion400m}. }
\label{app:tab:dataset_info}
\end{table}

%% file: tables/appendix/training.tex
\begin{table*}[t]
\centering
\resizebox{\linewidth}{!}{
\begin{tabular}{l|ccc}
\toprule
Config & Stage1 & Stage2 & Stage3 \\
\midrule
GPUs (V100) & \multicolumn{3}{c}{32 (Swin-T)/64 (Swin-L)} \\
training module & OV detector & object captioner & all modules \\
training objective & $\mathcal{L}_{det}=\mathcal{L}_{align}+\mathcal{L}_{box}+\mathcal{L}_{iou}$ & $\mathcal{L}_{lm}=\mathcal{L}_{lm}^{obj}+\mathcal{L}_{lm}^{img}$ & $\mathcal{L} = \mathcal{L}_{det} + \mathcal{L}_{lm}$  \\
training epochs & 12 & 3 & 5\\
input resolution & $320\times240\sim1333\times800$ & $320\times320$ & $1333\times600\sim1333\times800$\\
batch size  & 128 & 2048 & 128 \\
learning rate & 2.8e-4 & 1e-4 & 1e-4 \\
text encoder lr reduce factor & 0.1 & {--} &0.1\\
numper of concepts (grounding/image-text pairs) & 4800 & {--} & 4800 \\
optimizer & \multicolumn{3}{c}{AdamW \cite{adamw}} \\
optimizer momentum &  \multicolumn{3}{c}{$\beta_1=0.9$, $\beta_2=0.999$}\\
weight decay & \multicolumn{3}{c}{0.05} \\
warmup iters & \multicolumn{3}{c}{1000} \\
learning rate schedule & \multicolumn{3}{c}{cosine annealing} \\
text token length (text encoder) & \multicolumn{3}{c}{16} \\
text tokenizer (text encoder) & \multicolumn{3}{c}{CLIP~\cite{clip} Tokenizer} \\
text token length (object captioner) & \multicolumn{3}{c}{32} \\
text tokenizer (object captioner) & \multicolumn{3}{c}{BERT~\cite{bert} Tokenizer} \\
\bottomrule
\end{tabular}
}
\vspace{-.5em}
\caption{Detailed pre-training settings of \ourmodel. $\mathcal{L}_{lm}^{obj}$ and $\mathcal{L}_{lm}^{img}$ represent for language modeling training objective for object-level captioning and image-level captioning, respectively.}
\label{app:tab:training_detail}
\vspace{-.5em}
\end{table*}

%% file: tables/appendix/finetune_detail.tex
\begin{table}[t]
\centering
\resizebox{\linewidth}{!}{
\begin{tabular}{l|r}
\toprule
Config & Value \\
\midrule
GPUs (V100) & 16 \\
training epochs & 16 \\
optimizer & AdamW \cite{adamw} \\
optimizer momentum &  $\beta_1=0.9$, $\beta_2=0.999$ \\
lr for image encoder & 2e-5 \\
lr for text encoder & 2e-6 \\
weight decay & 0.05 \\
warmup iters & 1000 \\
learning rate schedule & cosine decay \\
batch size  & 64 \\
input resolution  & $1333\times800$\\
number of concepts per sample & 150\\
augmentation & \makecell[r]{ multi-scale training,\\ random flip} \\
\bottomrule
\end{tabular}
}
\vspace{-.5em}
\caption{Detailed fine-tuning settings for LVIS \cite{lvis}.}
\label{app:tab:finetune_lvis_details}
\vspace{-.5em}
\end{table}

\begin{table}[t]
\centering
\resizebox{\linewidth}{!}{
\begin{tabular}{l|r}
\toprule
Config & Value \\
\midrule
GPUs (V100) & 8\\
maximum training epochs & 250 \\
optimizer & AdamW \cite{adamw} \\
optimizer momentum &  $\beta_1=0.9$, $\beta_2=0.999$\\
lr for image encoder & 4e-5 \\
lr for text encoder & 4e-7 \\
weight decay & 0.05 \\
warmup iters & 500 \\
learning rate schedule & auto-step decay \\
lr decay tolerance $t_1$ (epochs) & 10 \\
training terminate tolerance $t_2$ (epochs) & 15 \\
minimum lr to stop decay & 1e-8 \\
batch size  & 32 \\
input resolution  & $1333\times800$\\
augmentation & \makecell[r]{ multi-scale training,\\ random flip}\\
\bottomrule
\end{tabular}
}
\vspace{-.5em}
\caption{Detailed fine-tuning settings for ODinW13 \cite{glip}.}
\label{app:tab:finetune_odinw_details}
\vspace{-1em}
\end{table}

%% file: tables/appendix/prompts.tex
\begin{enumerate}
    \item \textbf{Recaptioning with VLLM}: 
    We employ InstructBLIP~\cite{instructblip} to recaption 240K image-text pairs. To leverage information from the original caption texts, we use the following prompt: 
    
        \textit{``Given a noisy caption of the image: \{raw caption\}, write a detailed clean description of the image."}
    
    \item \textbf{Entity extraction using GPT-4}: 
    In this step, we first utilize GPT-4 to filter out non-entity descriptions from the captions generated by the VLLM. The prompt used is: 

        \textit{``Here is a caption for an image: \{caption\}. Extract the part of factual description related to what is directly observable in the image, while filtering out the parts that refer to inferred contents, description of atmosphere/appearance/style and introduction of history/culture/brand etc. Return solely the result without any other contents. If you think there is no factual description, just return `None'."}

    Subsequently, we extract information about object entities from the filtered captions using the prompt:

        \textit{``You are an AI tasked with developing an open-set object detection dataset from a large number of image captions, without access to the actual images. Your mission is to accurately identify and extract 'objects' from these captions, following the principles below:\\
        1. 'Objects' are physically tangible: They must be concrete entities that can be visually represented in an image. They are NOT (1) abstract  concepts (like 'history', 'culture') or feelings (like 'sorrow', 'happiness'), (2) meta-references to the image itself (e.g., 'image', 'picture', 'photo') or the camera (e.g. something is facing the 'camera'), unless they are specifically referring to physical elements within the image. (3) any descriptors (like 'appearance', 'atmosphere', 'color'), (4) events/activities and processes (like 'game', 'presentation', 'performance') and specific event types (like 'country style wedding', 'film festival'), (5) compositional aspects (like 'perspective', 'focus', 'composition') or viewpoint/perspective (like 'bird's eye view').\\
        2. 'Objects' are visually distinct: They are standalone entities that can be visually isolated from their environment. They do not include environmental characteristics (like 'colorful environment') and general location/scene descriptors (e.g., 'scene set indoors', 'country setting', 'sunny day', 'black and white illustration')\\
        Adhere to these guidelines for the extraction process:\\
        1. Consolidate duplicates: If multiple extracted 'objects' refer to the same entity in the caption, merge them into one while retaining conceptual diversity.\\
        2. Categorize the descriptive variants: For 'objects' described with adjectives, provide both versions - with and without the adjective.\\
        3. Identify the broader category: Assign a 'parent category' that each 'object' belongs to.\\
        Present your results as a numbered list in this format: id. 'object with adjective', 'object without adjective', 'parent category'. Your response should consist exclusively of results, with no superfluous content.\\
        Here's the caption: \{caption\}"}
    
    \item \textbf{Instruction tuning of VLLM for large-scale annotation}: 
    In this phase, we use the caption texts and object entity information obtained from the above steps to fine-tune the LLaVA~\cite{llava} model. Here, we combine the aforementioned information into a new concise prompt, and the question-answer pair is constructed as:
    
         \textit{\textbf{Question}: \\ ``From the noisy caption of the image: \{raw caption\}, generate a refined image description and identify all visible 'objects' -- any visually and physically identifiable entity in the image. Keep the following guidelines in mind:\\
            1. Merge similar 'objects' from the caption, preserving conceptual diversity.\\
            2. For adjective-described 'objects', provide versions both with and without the adjective.\\
            3. Assign a 'parent category' for each 'object'.\\
            Present results as:\\
            Caption: \{caption\}\\
            Objects: \{id. 'object with adjective', 'object without adjective', 'parent category'\}.\\
            $<$image tokens$>$"}

         \textit{\textbf{Answer}: \\Caption: \{refined caption\}\\
            Objects: \{entity information\}}

Here, the VLLM receives image tokens, \ie, $<$image tokens$>$, along with their original captions, \ie, \textit{\{raw caption\}}, as inputs, and learns to generate refined captions and extract information about object entities.
\end{enumerate}

%% file: tables/appendix/lvis.tex
\begin{table*}[t]
\vspace{-4mm}
\centering
\tablestyle{3pt}{1}
\resizebox{\linewidth}{!}{
\begin{tabular}{cllcccccccccc}
\toprule
\multirow{2}{*}{}&\multirow{2}{*}{Method} & \multirow{2}{*}{Backbone} & Pre-training & Fine-tuning  &  \multicolumn{4}{c}{$\text{LVIS}^{\text{minival}}$} &\multicolumn{4}{c}{$\text{LVIS}^{\text{val}}$} \tabularnewline
& &  & data & data & $\text{AP}_{\text{all}}$ & $\text{AP}_{\text{r}}$ & $\text{AP}_{\text{c}}$ & $\text{AP}_{\text{f}}$ &$\text{AP}_{\text{all}}$ & $\text{AP}_{\text{r}}$ & $\text{AP}_{\text{c}}$ & $\text{AP}_{\text{f}}$\\
 
\midrule

1 & OWL-ST~\cite{scaleovdet} & CLIP B/16 & WebLI2B & {--} & {31.8} & {35.4} & {--} & {--}  &{27.0}&{29.6}&{--}&{--} \\
2 & OWL-ST~\cite{scaleovdet} & CLIP L/14 & WebLI2B & {--}  & {38.1} & {39.0} & {--} & {--}  &{33.5}&{34.9}&{--}&{--} \\
\rowcolor{lightblue} 3 &  \ourmodel  & Swin-T & {O365,V3Det,GoldG,\ourdata} & {--} & {43.7} & {39.3} & {44.5} & {43.7} &{36.7}&{34.2}&{34.9}&{39.9} \\
\rowcolor{lightblue} 4 & \ourmodel  & Swin-L & {O365,V3Det,GoldG,\ourdata} & {--}  & {45.8} & {46.9} & {45.9} & {45.5}&{39.6}&{38.9}&{38.4}&{41.3} \\
\midrule
5 & OWL-ST+FT~\cite{scaleovdet} & CLIP B/16 & WebLI2B & $\text{LVIS}_\text{base}$  & {47.2} & {37.8} & {--} & {--}  &{41.8} & {36.2} &{--}&{--} \\
6 & OWL-ST+FT~\cite{scaleovdet} & CLIP L/14 & WebLI2B & $\text{LVIS}_\text{base}$  & {54.1} & {46.1} & {--} & {--}  &{49.4} & {44.6} &{--}&{--} \\
\rowcolor{lightblue} 7 &  \ourmodelft  & Swin-T & {O365,V3Det,GoldG,\ourdata} & $\text{LVIS}_\text{base}$ & {54.4} & {46.7} & {56.1} & {54.3} &{48.2}&{40.2}&{48.5}&{51.3} \\
\rowcolor{lightblue} 8 & \ourmodelft  & Swin-L & {O365,V3Det,GoldG,\ourdata} & $\text{LVIS}_\text{base}$ & {60.8} & {56.7} & {63.2} & {59.4}&{54.1}&{45.8}&{55.4}&{56.4} \\
\bottomrule
\end{tabular}
}
\vspace{-.7em}
\caption{Zero-shot and fine-tuning AP on LVIS val~\cite{lvis}  and minival~\cite{mdetr}. Results labeled without `+FT' represent zero-shot performance,  whereas those with '+FT' indicate results of fine-tuning with LVIS base categories ($\text{LVIS}_\text{base}$). \ourmodel significantly 
 outperforms OWL-ST~\cite{scaleovdet}, which is pre-trained with 2 billion image-text pairs.}
\label{app:tab:lvis} \vspace{-1.2em}
\end{table*}

%% file: tables/appendix/coco-o.tex
\begin{table*}[t]
    \centering
    \centerline{
    \resizebox{0.8\textwidth}{!}{%
    \begin{tabular}{lcccccccc}
    \toprule
    Method & Backbone &  {Sketch} &	Weather &	Cartoon	 & Painting	& Tattoo	& Handmake	& Average \\
    \midrule
    \ourmodel & {Swin-T} & 38.3 & 43.6 & 45.0 & 43.2 & 29.3 & 31.5 & 38.5 \\
    \ourmodel & {Swin-L} & 50.8 & 48.6 & 56.9 & 53.7 & 44.5 & 38.2 & 48.8 \\
    \bottomrule
    \end{tabular}
    }
    }
    \vspace{-.7em}
    \caption{Detailed performance on COCO-O~\cite{coco-o} dataset. Zero-shot AP is reported.}
    \vspace{-0.5em}
    \label{app:tab:coco-o_full_results}
\end{table*}

%% file: tables/appendix/odinw.tex
\newcommand*\rot{\rotatebox{90}}
\begin{table*}[t]
    \centering
    \centerline{
    \resizebox{1.0\textwidth}{!}{%
    \vspace{-1.5em}
    \begin{tabular}{clccccccccccccccc}
    \toprule
    &\rot{Method} & \rot{Backbone} & \rot{PascalVOC} & \rot{AerialDrone} & \rot{Aquarium} & \rot{Rabbits} & \rot{EgoHands} & \rot{Mushrooms} & \rot{Packages} & \rot{Raccoon} & \rot{Shellfish} & \rot{Vehicles} & \rot{Pistols}& \rot{Pothole}  & \rot{Thermal} & \rot{\textbf{Average}} \\
    \midrule
    1 & GLIP~\cite{glip} & {Swin-T} & 62.3 & 31.2 & 52.5 & 70.8 & 78.7 & 88.1 & 75.6 & 61.4 & 51.4 & 65.3 & 71.2 & 58.7 & 76.7 & 64.9 \\
    2 & GLIPv2~\cite{glipv2} & {Swin-T} & 66.4 & 30.2 & 52.5 & 74.8 & 80.0 & 88.1 & 74.3 & 63.7 & 54.4 & 63.0 & 73.0 & 60.1 & 83.5 & 66.5 \\
    3 & GLIPv2~\cite{glipv2} & {Swin-B} & 71.1 & 32.6 & 57.5 & 73.6 & 80.0 & 88.1 & 74.9 & 68.2 & 70.6 & 71.2 & 76.5 & 58.7 & 79.6 & 69.4 \\
    4 & DetCLIPv2~\cite{detclipv2} & {Swin-T} & 67.5 & 41.8 & 50.8 & 80.4 & 79.8 & 90.1 & 73.7 & 70.8 & 54.8 & 66.5 & 77.7 & 54.8 & 82.2 & 68.5 \\
    \rowcolor{lightblue} 5 & \ourmodel & {Swin-T} & 72.5 & 51.6 & 54.5 & 79.9 & 81.2 & 94.1 & 78.2 & 71.6 & 53.9 & 67.4 & 79.4 & 55.1 & 84.4 & \textbf{71.1}
\\
    \midrule
    6 & GLIP~\cite{glip} & {Swin-L} & 69.6 & 32.6 & 56.6 & 76.4 & 79.4 & 88.1 & 67.1 & 69.4 & 65.8 & 71.6 & 75.7 & 60.3 & 83.1 & 68.9 \\
    7 & GLIPv2~\cite{glipv2} & {Swin-H} & 74.4 & 36.3 & 58.7 & 77.1 & 79.3 & 88.1 & 74.3 & 73.1 & 70.0 & 72.2 & 72.5 & 58.3 & 81.4 & 70.4 \\
    8 & DetCLIPv2~\cite{detclipv2} & {Swin-L} & 74.4 & 44.1 & 54.7 & 80.9 & 79.9 & 90 & 74.1 & 69.4 & 61.2 & 68.1 & 80.3 & 57.1 & 81.1 & 70.4 \\
    \rowcolor{lightblue} 9 & \ourmodel & {Swin-L} & 76.4 & 51.2 & 57.5 & 79.9 & 80.2 & 90.4 & 75.1 & 70.9 & 63.6 & 69.8 & 82.7 & 56.2 & 83.8 & \textbf{72.1} \\
    \bottomrule
    \end{tabular}
    }
    }
    \vspace{-.7em}
    \caption{Detailed fine-tuned AP on ODinW13~\cite{glip} dataset. \ourmodel outperforms its counterparts by a large margin. }
    \vspace{-.7em}
    \label{app:tab:odinw-full-results}
\end{table*}

%% file: tables/appendix/inference_speed.tex
\begin{table}
\begin{tabular}{llcc}
\toprule
Method & Backbone & OV detector & \makecell{Object \\captioner} \\
\midrule
GLIP~\cite{glip}& Swin-T & 2.5 FPS & {--} \\
DetCLIP~\cite{detclip} &   Swin-T & 2.3 FPS & {--} \\
\rowcolor{lightblue} \ourmodel & Swin-T & \textbf{14.5} FPS & 1.2 FPS \\
\midrule
GLIP~\cite{glip} & Swin-L & 0.3 FPS & {--} \\
\rowcolor{lightblue} \ourmodel & Swin-L & \textbf{8.2} FPS & 0.9 FPS \\

\bottomrule
\end{tabular}
\vspace{-.7em}
\caption{Inference speed. We test the speed with V100 GPU, using batch size=1 and FP16 inference. \ourmodel can run significantly faster than previous methods like GLIP~\cite{glip} and DetCLIP~\cite{detclip}. }
\label{app:tab:inference speed} 
\vspace{-1.2em}
\end{table}